\documentclass[10pt,twocolumn,letterpaper]{article}

\usepackage{cvpr}              % To produce the CAMERA-READY version
\usepackage[dvipsnames]{xcolor}

\usepackage[accsupp]{axessibility}
\usepackage{graphicx}
\usepackage{amsmath}
\usepackage{amssymb}
\usepackage{booktabs}
\usepackage{xcolor}
\usepackage{colortbl}
\usepackage{bbding}
\usepackage{wrapfig}
\usepackage{setspace}
\usepackage{graphicx}       % to insert images
\usepackage{multirow}
\usepackage{color}
\usepackage{kotex}
\usepackage{kotex-logo}
\usepackage{algorithm}
\usepackage{algpseudocode}
\usepackage{arydshln}
\usepackage{cite}

\usepackage{caption}
\usepackage{adjustbox}
\usepackage{wrapfig}
\usepackage{lipsum}
\usepackage{array}
\usepackage{makecell}

\definecolor{cvprblue}{rgb}{0.21,0.49,0.74}
\usepackage[pagebackref,breaklinks,colorlinks,citecolor=cvprblue]{hyperref}

\usepackage[capitalize]{cleveref}
\crefname{section}{Sec.}{Secs.}
\Crefname{section}{Section}{Sections}
\Crefname{table}{Table}{Tables}
\crefname{table}{Tab.}{Tabs.}

\def\cvprPaperID{8753} % *** Enter the CVPR Paper ID here
\def\confName{CVPR}
\def\confYear{2024}

\begin{document}

 \author{Seokju Yun\qquad Youngmin Ro\thanks{Corresponding author (E-mail: \tt youngmin.ro@uos.ac.kr)}\\Machine Intelligence Laboratory, University of Seoul, Korea \\ %\tt\small \{wsz871, youngmin.ro\}@uos.ac.kr \\
\tt \small Code: \href{https://github.com/ysj9909/SHViT}{https://github.com/ysj9909/SHViT}}

% \title{\LaTeX\ Author Guidelines for \confName~Proceedings}
\title{SHViT: Single-Head Vision Transformer with Memory Efficient Macro Design}

%\author[1]{Seokju Yun}
%\author[2]{Youngmin Ro}

%\author{Seokju Yun \and Youngmin Ro\\ 
%Machine Intelligence Laboratory, University of Seoul, Korea\\
%\tt\small \{wsz871, youngmin.ro\}@uos.ac.kr
%}

% Additional authors and addresses can stand added with ``\and'',
% \and
% Youngmin Ro\\
% % \Machine Intelligence Laboratory, University of Seoul, Korea \\
% {\tt\small youngmin.ro@uos.ac.kr}

\maketitle

\begin{abstract} 
Recently, efficient Vision Transformers have shown great performance with low latency on resource-constrained devices.
Conventionally, they use 4$\times$4 patch embeddings and a 4-stage structure at the macro level, while utilizing sophisticated attention with multi-head configuration at the micro level.
This paper aims to address computational redundancy at all design levels in a memory-efficient manner.
We discover that using larger-stride patchify stem not only reduces memory access costs but also achieves competitive performance by leveraging token representations with reduced spatial redundancy from the early stages.
Furthermore, our preliminary analyses suggest that attention layers in the early stages can be substituted with convolutions, and several attention heads in the latter stages are computationally redundant.
To handle this, we introduce a single-head attention module that inherently prevents head redundancy and simultaneously boosts accuracy by parallelly combining global and local information.
Building upon our solutions, we introduce SHViT, a Single-Head Vision Transformer that obtains the state-of-the-art speed-accuracy tradeoff.
For example, on ImageNet-1k, our SHViT-S4 is 3.3$\times$, 8.1$\times$, and 2.4$\times$ faster than MobileViTv2 $\times$1.0 on GPU, CPU, and iPhone12 mobile device, respectively, while being 1.3\% more accurate.
For object detection and instance segmentation on MS COCO using Mask-RCNN head, our model achieves performance comparable to FastViT-SA12 while exhibiting 3.8$\times$ and 2.0$\times$ lower backbone latency on GPU and mobile device, respectively.

\end{abstract}
\vspace{-10pt}
\section{Introduction}\label{intro}
Vision Transformers (ViT) have demonstrated impressive performance across various computer vision tasks due to their high model capabilities~\cite{vit, detr, maskformer}.
Compared to Convolutional Neural Networks (CNN)~\cite{he2016resnet, liu2022convnext}, ViTs excel in modeling long-range dependencies and scale effectively with large amounts of training data and model parameters~\cite{raghu2021vitdocnn}.
Despite these advantages, the lack of inductive bias in vanilla ViTs necessitates more training data, and the global attention module incurs quadratic computational complexity with respect to the image size. 
To address these challenges, previous research has either combined ViTs with CNNs or introduced cost-efficient attention variants.

Recently, studies addressing problems with real-time constraints have also proposed efficient models following similar strategies.
And their strategies can be categorized into two groups: 1) efficient architecture - \textit{macro design}; and 2) efficient Multi-Head Self-Attention (MHSA) - \textit{micro design}.
Studies exploring architectural design~\cite{vasu2023fastvit, li2022efficientformer, li2022efficientformerv2, graham2021levit, maaz2023edgenext, alternet} utilize convolution to handle high-resolution / low-level features and employ attention for low-resolution / high-level features, demonstrating superior performance without complex operations.
%Research focusing  on architectural design involves amalgamation of ViTs and CNNs to build efficient structures~\cite{mobileformer, EMO, vasu2023fastvit, li2022efficientformer, li2022efficientformerv2, graham2021levit, maaz2023edgenext, alternet}. 
\begin{figure}[t]
    \centering
    
    % 이미지 부분
    \includegraphics[width=1.0\linewidth]{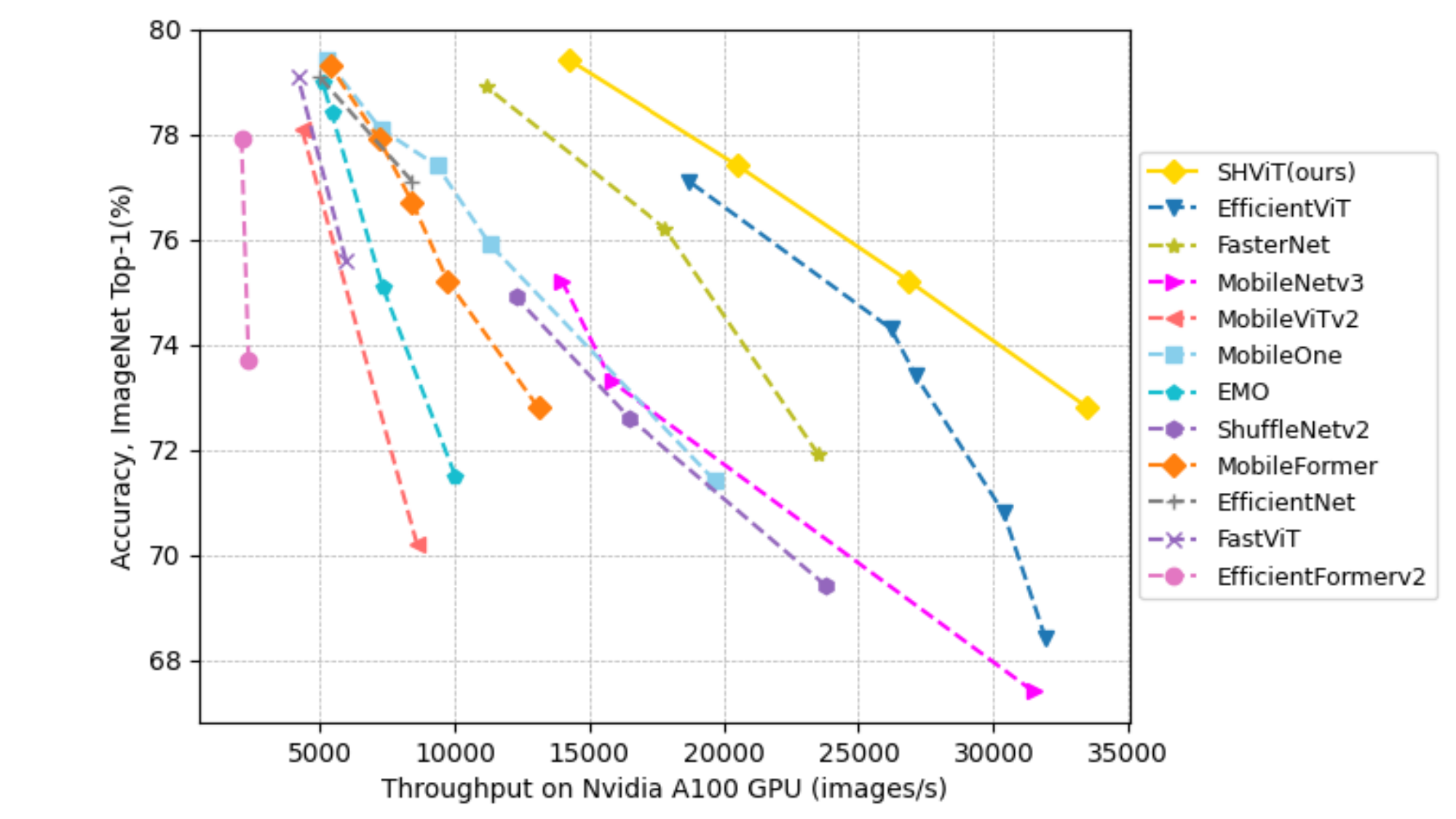}
    \vspace{-20pt}
    \caption{Comparison of throughput and accuracy between our SHViT and other recent methods.}
    \vspace{-10pt}
    \label{fig: acc_fps_plot}
\end{figure}
%Comparison of throughput and accuracy between SHViT and other   r recent methods.
%Accuracy \textit{vs.} Throughput with recent methods.
However, most of these methods mainly focus on which modules to use for aggregating tokens rather than how to construct the tokens (about patchify stem and stage design).
On the other hand, efficient MHSA techniques reduce the cost of attention by implementing sparse attention~\cite{kitaev2020reformer, pan2022hilo, ren2021combiner, wang2021pyramid, pan2022edgevits, li2022efficientformerv2, mehta2022mobilevit} or low-rank approximation~\cite{mobileformer, choromanski2020performer, mehta2022mobilevitv2, wang2020linformer}.
These modules are applied with the commonly adopted multi-head mechanism.
Despite all the great progress, redundancies in macro/micro design are still not fully understood or addressed.
\textit{In this paper, we explore the redundancy at all design levels, and propose memory-efficient solutions.}

To identify computational redundancy in macro design, we concentrate on patch embedding size, observing that most recent efficient models use a 4$\times$4 patch embedding.
We conduct experiments as shown in Fig.~\ref{fig: analysis_macro} to analyze spatial redundancy in traditional macro design and find several intriguing results.
First, despite having fewer channels, the early stages exhibit a severe speed bottleneck due to the large number of tokens (at 224$\times$224, stage1: 3136 tokens; stage2: 784 tokens).
Second, using a 3-stage design that processes 196 tokens in the first stage through a 16$\times$16 patchify stem does not lead to a significant drop in performance.
%Inspired by ~\cite{graham2021levit, xiao2021early, liu2023efficientvit}, we hypothesize that a 16$\times$16 overlapping patch embedding has two distinct advantages (it will be analyzed in sec.~\ref{sec: macro_design}).
%First, it leverages a broader receptive field of low-level visual representations, minimizing spatial redundancy and thereby enhancing the model's capacity.
%Second, it boosts inference speed by reducing the memory access cost of feature maps by up to 16 times.
For further comparison, we set up a basic model (Tab.~\ref{tab: ablation} (2)) employing the aforementioned macro design, performing simple token mixing using a 3$\times$3 depthwise convolution.
Compared to the efficient model MobileViT-XS~\cite{mehta2022mobilevit}, our simple model achieves 1.5\% superior accuracy on ImageNet-1k~\cite{deng2009imagenet}, while running \textbf{5.0$\times$ / 7.6$\times$} faster on the A100 GPU / Intel CPU.
\textit{These results demonstrate that there is considerable spatial redundancy in the early stages, and compared with specialized attention methods, efficient macro design is more crucial for the model to achieve competitive performance within strict latency limits.}
Note that this observation does not mean the token mixer is trivial.

We also probe redundancy in micro design, specifically within the MHSA layer.
Most efficient MHSA methods have primarily focused on effective spatial token mixing.
Due to the efficient macro design, we are able to use compact token representations with increased semantic density. 
%Therefore, we focus on channel(head) redundancy present in attention layers. 
Thus, we turn our focus to the channel (head) redundancy present in attention layers, also crucial aspect overlooked in most previous works.
Through comprehensive experiments, we find that there is a noticeable redundancy in multi-head mechanism, particularly in the latter stages.
We then propose a novel Single-Head Self-Attention (SHSA) as a competitive alternative that reduces the computational redundancy.
In SHSA, self-attention with a single head is applied to just a subset of the input channels, while the others remain unchanged.
%It simply applies a self-attention on only a part of the input channels with single head, leaving the remaining channels untouched.
\textit{SHSA layer not only eliminates the computational redundancy derived from multi-head mechanism but also reduces memory access cost by processing partial channels.}
Also, these efficiencies enable stacking more blocks with a larger width, leading to performance improvement within the same computational budget.

Based on these findings, we introduce a Single-Head Vision Transformer (SHViT) based on memory-efficient design principles, as a new family of networks that run highly fast on diverse devices.
Experiments demonstrate that our SHViT achieves state-of-the-art performance for classification, detection, and segmentation tasks in terms of both speed and accuracy, as shown in Fig.~\ref{fig: acc_fps_plot}.
For instance, our SHViT-S4 achieves 79.4\% top-1 accuracy on ImageNet with throughput of 14283 images/s on an Nvidia A100 GPU and 509 images/s on an Intel Xeon Gold 5218R CPU @ 2.10GHz, outperforming EfficientNet-B0~\cite{efficientnet} by 2.3\% in accuracy, 69.4\% in GPU inference speed, and 90.6\% in CPU speed.
Also, SHViT-S4 has 1.3\% better accuracy than MobileViTv2$\times$1.0~\cite{mehta2022mobilevitv2} and is 2.4$\times$ faster on iPhone12 mobile device.
For object detection and instance segmentation on MS COCO~\cite{lin2014microsoftcoco} using Mask-RCNN~\cite{he2017mask} detector, our model significantly outperforms EfficientViT-M4~\cite{liu2023efficientvit} by 6.2 AP$^{box}$ and 4.9 AP$^{mask}$ with a smaller backbone latency on various devices.

\noindent In summary, our contributions are as follows:
\begin{itemize}
    \item We conduct a systematic analysis of the redundancy that has been overlooked in the majority of existing research, and propose memory-efficient design principles to tackle it.
    \item We introduce Single-Head ViT(SHViT), which strike a good accuracy-speed tradeoff on a variety of devices such as GPU, CPU, and iPhone mobile device.
    \item We carry out extensive experiments on various tasks
and validate the high speed and effectiveness of our SHViT.
\end{itemize}

\definecolor{m}{RGB}{230, 230, 230}
\section{Analysis and Method}
In this section, we first conduct analyses of redundancies in both macro and micro design through first-of-its-kind experiments and then discuss various solutions to mitigate them. 
%Moreover, we provide an in-depth analysis of the advantages inherent to our proposed methodologies.
After that, we introduce the Single-Head Vision Transformer (SHViT) and explain its details.

\colorlet{lightblue}{blue!50}
\colorlet{lightred}{red!50}
\newcolumntype{C}[1]{>{\centering\arraybackslash}p{#1}}
\begin{figure}[t]
    \centering
    \large
    
    % 이미지 부분
    %$\includegraphics[height=3.5cm,width=1.0\linewidth]{Fig/analysis_macro_.pdf}
    \includegraphics[width=1.0\linewidth]{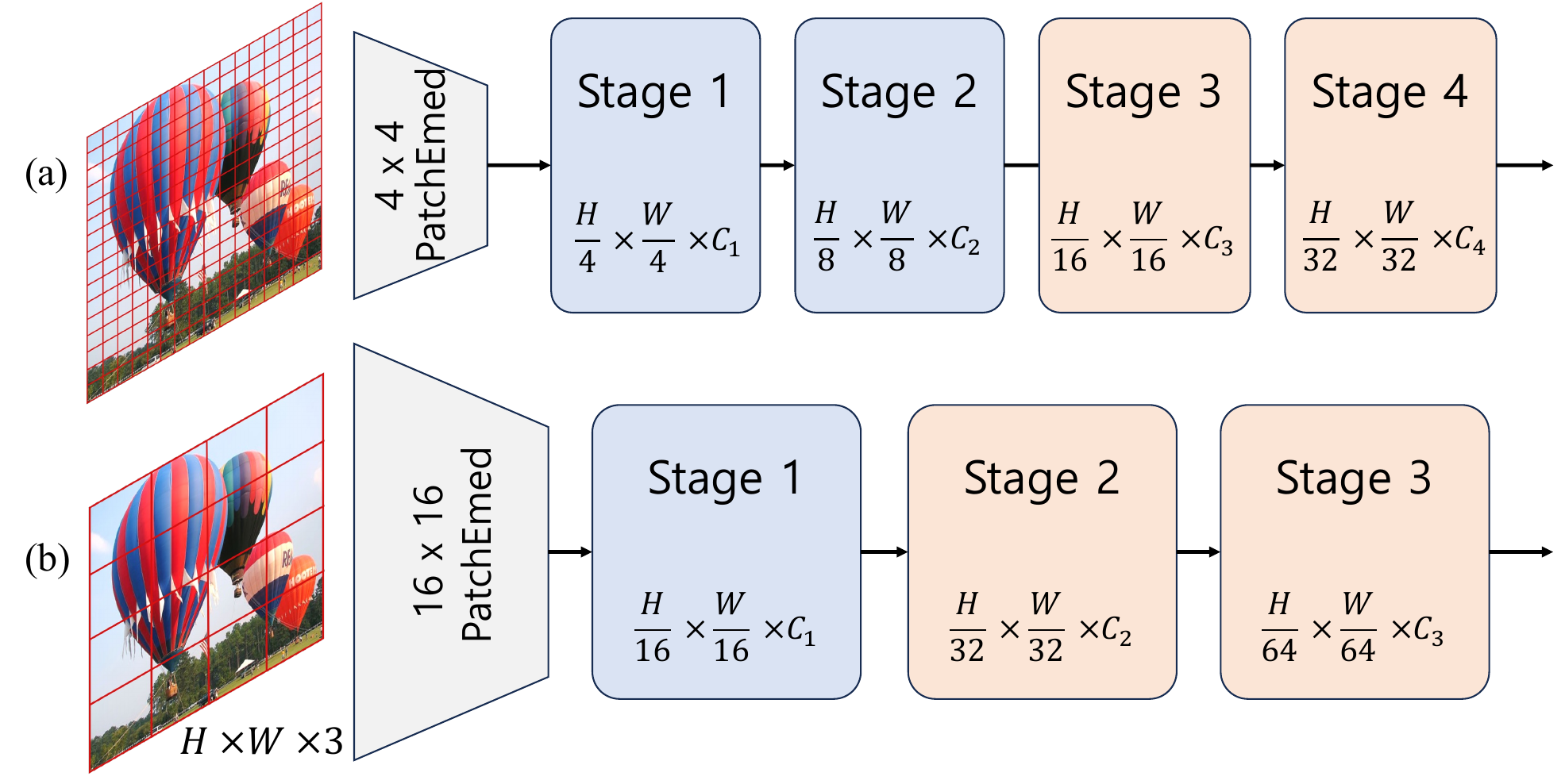}

    %\caption{My Image}
    
    %\vspace{1cm} % 이미지와 테이블 사이에 약간의 공간을 추가
    
    % 테이블 부분
    \resizebox{1.0\linewidth}{!}{%
    \begin{tabular}{lC{1.0cm}C{1.5cm}C{2.5cm}C{2.5cm}C{2cm}}
    \hline
    Macro design &  Reso. & \; Params \newline (M) & CPU$_{\text{ONNX}}$\newline \text{(images/s)} & \; \, GPU \newline (images/s) & \; Top-1 \newline (\%) \\
    \hline
    (a) : [\textcolor{lightblue}{24(1)}, \textcolor{lightblue}{48(1)}, \textcolor{lightred}{184(4)}, \textcolor{lightred}{368(2)}] & 224 & 4.9 & 430 & 11267 & 75.0 \\
    (b) : [\textcolor{lightblue}{128(2)}, \textcolor{lightred}{256(4)}, \textcolor{lightred}{384(2)}] & 224 & 7.2 & 1183 \small \textcolor{red}{(2.8× $\uparrow$)} & 33810 \small \textcolor{red}{(3.0× $\uparrow$)} & 73.5 \small \textcolor{blue}{(-1.5)} \\
    (b$'$) : [\textcolor{lightblue}{128(2)}, \textcolor{lightred}{256(4)}, \textcolor{lightred}{384(2)}] & 256 & 7.2 & 1064 \small \textcolor{red}{(2.5× $\uparrow$)} & 27641 \small \textcolor{red}{(2.5× $\uparrow$)} & 74.9 \small \textcolor{blue}{(-0.1)} \\
    \hline
    (b$''$) : [\textcolor{lightred}{128(2)}, \textcolor{lightred}{256(4)}, \textcolor{lightred}{384(2)}] & 224 & 7.3 & 729 & 26101 & 74.0 \\
    \hline
    \end{tabular} }
    \caption{\textbf{Macro design analysis.} All stages are composed of MetaFormer blocks~\cite{yu2022metaformer}. The stages depicted in blue and red utilize depthwise convolution and attention layers as token mixer, respectively. In the table below, the macro design numbers represent the number of channels, while the numbers in parentheses indicate the number of blocks.}
    \label{fig: analysis_macro}
\end{figure}

\subsection{Analysis of Redundancy in Macro Design}\label{sec: macro_design}
Most efficient models~\cite{pan2022edgevits, EMO, maaz2023edgenext, chen2023fasternet, mehta2022mobilevitv2, vasu2023fastvit, li2022efficientformer, li2022efficientformerv2, mobileformer, liu2022convnext} adopt a 4$\times$4 patchify stem / 4-stage configuration (Fig.~\ref{fig: analysis_macro} (a)).
In contrast, plain ViT models~\cite{vit, deit} adopt a 16$\times$16 patchify stem to generate meaningful input tokens for subsequent MHSA layers.
We focus on this discrepancy and further hypothesize that a larger-stride patchify stem is not only necessary for the MHSA layers but also crucial for effective representation learning within tight latency regimes.

To substantiate our hypothesis, inspired by ~\cite{graham2021levit, xiao2021early, liu2023efficientvit}, we adopt 16$\times$16 patchify stem and 3-stage design.
We build two models based on the MetaFormer block~\cite{yu2022metaformer} and the two aforementioned macro designs (see Fig.~\ref{fig: analysis_macro} for details).
Specifically, we configured both models to have a similar number of channels for equivalent feature map size.
Surprisingly, model (b) is 3.0$\times$ / 2.8$\times$ faster on the GPU / CPU, respectively, although it performs 1.5\% worse than (a).
Furthermore, when trained at a resolution of 256$\times$256, (b$'$) is not only comparable to (a) but also significantly faster.

As shown in the above observations, our proposed efficient macro design has the following advantages: 

\noindent \textbf{1)} token representations with large receptive fields and reduced spatial redundancy can be utilized at the early stages. \textbf{2)} It can diminish the feature map size by up to 16 times, leading to a significant reduction in memory access costs. \textbf{3)} due to the aggressive stride design, there's only a mild decrease in throughput when the resolution is increased, leading to effective performance enhancement (as shown in Fig.~\ref{fig: analysis_macro}(b$'$),~\ref{fig: mobile_reso}, and Tab.~\ref{tab: imagenet_results}).

\begin{figure}[t]
    \centering
    
    % 이미지 부분
    \includegraphics[width=1.0\linewidth]{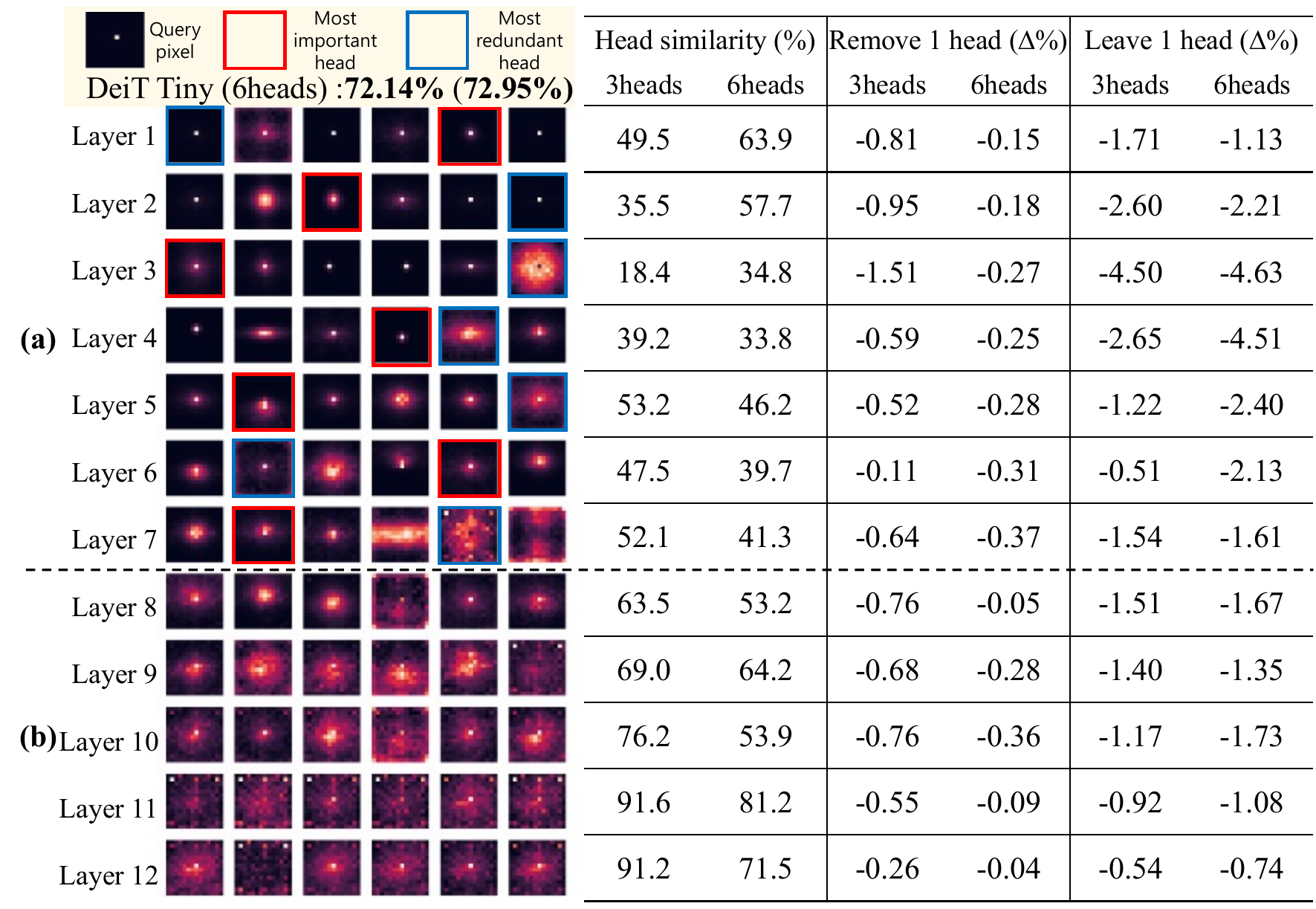}
    \caption{\textbf{Multi-head redundancy analysis on DeiT~\cite{deit}.} To better analyze head redundancy, we increase the number of heads in DeiT-T from 3 to 6 and retrain the model. We compute the attention maps and calculate the average cosine similarity between each head in different layers across 128 test samples from ImageNet. The importance of each head is determined by its score when it is removed and when it is left alone. 
    %Results in parentheses are from the orginal DeiT-T (3 heads).
    Zoom-in for better visibility.}
    \label{fig: deit_redundancy}
\end{figure}

\subsection{Analysis of Redundancy in Micro Design}\label{sec: micro_design}
MHSA layer computes and applies attention maps independently in multiple subspaces (heads), which has consistently shown enhanced performance~\cite{vit, attention}.
However, while attention maps are computationally demanding, recent studies have shown that many of them aren't critically essential~\cite{michel2019sixteen, voita2019analyzing, chen2021svite, song2022cpvit, hou2022multicompression, yang2023vitprune}.
We also delve into the multi-head redundancy of prevailing tiny ViT models (DeiT-T~\cite{deit}, Swin-T~\cite{liu2021swin}) through three experiments:\textit{attention map visualization, head similarity analysis, and head ablation study.}

\noindent For head similarity analysis, we measure the average cosine similarity between each head and other heads in the same layer.
For head ablation study, we evaluate the performance impact by nullifying the output of some heads in a given layer while maintaining full heads in the other layers.
And the highest score is reported.
Details of each experiment and further results are provided in the supplementary materials.

First of all, in the early stages (Fig.~\ref{fig: deit_redundancy} (a)), we observe that the top-performing heads tend to operate in a convolution-like manner, while heads that have minimal impact on performance when removed typically process information more globally.
Also, as shown in Fig.~\ref{fig: analysis_macro} (b$''$), the model using attention layers in the first stage exhibits a less favorable speed-accuracy trade-off compared to those employing depthwise convolution layers in the first stage.
Hence, for efficiency, we use convolutions with spatial inductive bias as the token mixer in the initial stage.

\begin{figure}[t]
    \centering
    
    % 이미지 부분
    \includegraphics[width=1.0\linewidth]{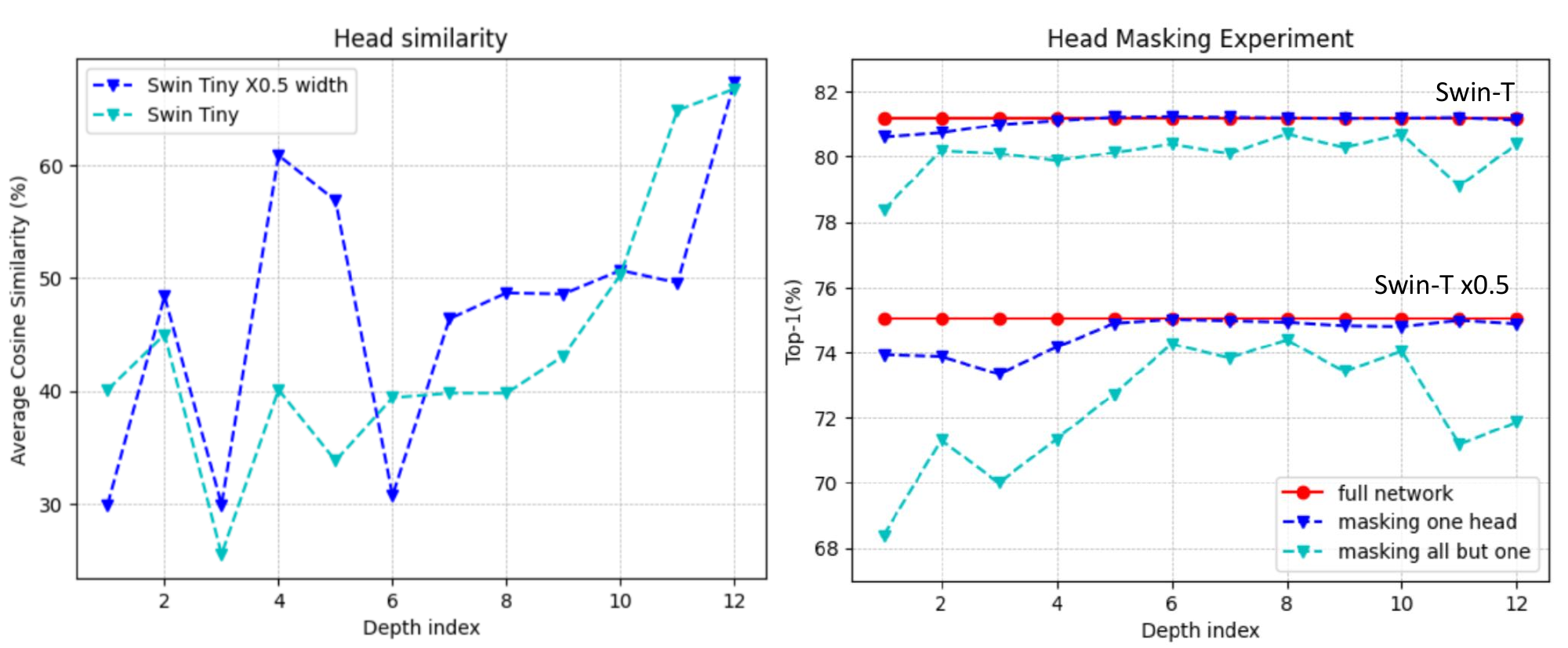}
    \vspace{-20pt}
    \caption{\textbf{Multi-head redundancy analysis on Swin~\cite{liu2021swin}.} We scale down by halving the width of Swin-T. \textbf{Left:} the average cosine similarity. \textbf{Right:} head masking results. The process of deriving the results is the same as the DeiT experiment. 
(Fig.~\ref{fig: deit_redundancy})}
    \label{fig: swin_redundancy}
\end{figure}

In the latter stages, we find that there is a lot of redundancy both at the feature and prediction levels.
%, as shown in Fig.~\ref{fig: deit_redundancy},~\ref{fig: swin_redundancy}.
For example, the latter stages of DeiT-T (Fig.~\ref{fig: deit_redundancy} (b)) exhibit the average head similarity of 78.3\% (64.8\% for 6 heads), with Swin-T also demonstrating notably high values (Fig.~\ref{fig: swin_redundancy} Left).
In the experiment of removing one head, We observe that the majority of heads can be removed without deviating too much from the original accuracy.
Remarkably, in some cases of Swin-T (Fig.~\ref{fig: swin_redundancy} Right), removing a head even leads to slightly improved score.
Furthermore, when using just one head out of 12 or 24 in Swin-T, the performance drop is, on average, only 0.95\% points.

Previous approaches~\cite{michel2019sixteen, voita2019analyzing, chen2021svite, song2022cpvit, hou2022multicompression, yang2023vitprune} to tackle head redundancy typically train full networks first and then prune unnecessary heads.
\begin{figure*}[t]
    \centering
    \includegraphics[width=0.9\linewidth]{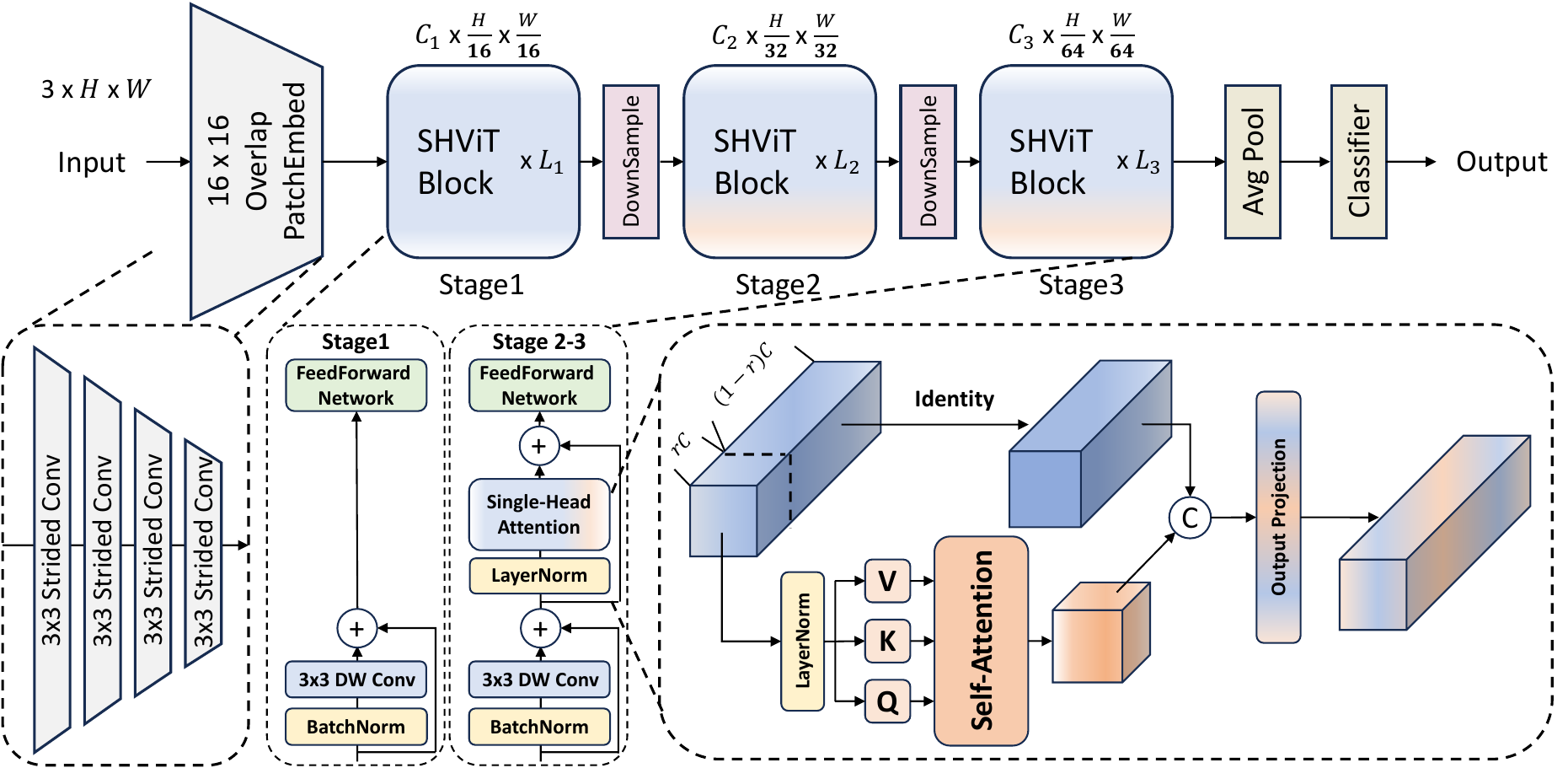}
    \caption{\textbf{Overview of Single-Head Vision Transformer (SHViT).} The model starts with a 16$\times$16 overlapping patch embedding layer and uses single-head attention layers in the latter stages to efficiently compute global dependencies. See text for details.}
    \label{fig:Model}
\end{figure*}
Although these methods are effective, they come at the expense of increased computational resources and memory footprints during training.
%To address the above problem in a cost-efficient manner, we propose Single-Head Self-Attention (SHSA) to block head redundancy at the architecture level.
%To address the above problem in a cost-efficient manner, we design our attention module with single head to  inherently avoid head redundancy, ensuring both training and inference processes are streamlined and efficient.
To address the aforementioned problem cost-effectively, we design our attention module with a single head to inherently avoid head redundancy.
This approach ensures both the training and inference processes are streamlined and efficient.

\subsection{Single-Head Self-Attention}\label{SHSA}
%To effectively handle attention features with a single head, we explore various architectural designs and   
%Results in Figure~\ref{fig: single_head_choices}
Based on the above analyses, we propose a new Single-Head Self-Attention (SHSA), with details presented in the lower right corner of Fig.~\ref{fig:Model}.
It simply applies an attention layer with a single head on only a part of the
input channels ($C_{p}$=$rC$) for spatial feature aggregation and leaves the remaining channels untouched.
We set $r$ to $1/4.67$ as a default.
Formally, the SHSA layer can be described as:
\begin{gather}
    \mathrm{SHSA}(\textbf{X}) = \mathrm{Concat}(\mathbf{\Tilde{X}}_{att}, \mathbf{X}_{res})W^O \\
    \mathrm{\Tilde{X}}_{att} = \mathrm{Attention}(\textbf{X}_{att}W^Q, \textbf{X}_{att}W^K, \textbf{X}_{att}W^V), \\
    \mathrm{Attention}(\textbf{Q}, \textbf{K}, \textbf{V}) = \mathrm{Softmax}(\textbf{Q}\textbf{K}^\mathsf{T} / \sqrt{d_{qk}})\textbf{V}, \\
    \mathrm{X}_{att}, \mathrm{X}_{res} = \mathrm{Split}(\textbf{X}, [C_p, C - C_p])
\end{gather}
\noindent where $W^Q$, $W^K$, $W^V$, and $W^O$ are projection weights, $d_{qk}$ is the dimension of the query and key (set to 16 as a default), and $\mathrm{Concat}(·)$ is the concatenation operation. 
For consistent memory access, we take the initial $C_{p}$ channels as representatives of the whole feature maps.
Additionally, the final projection of SHSA is applied to all channels, rather than just the initial $C_{p}$ channels, ensuring efficient propagation of the attention features to the remaining channels.
\textit{SHSA can be interpreted as sequentially stretching the previously parallel-computed redundant heads along the block-axis.}

\begin{figure}[t]
    \centering
    \large
    
    % 이미지 부분
    \includegraphics[width=1.0\linewidth]{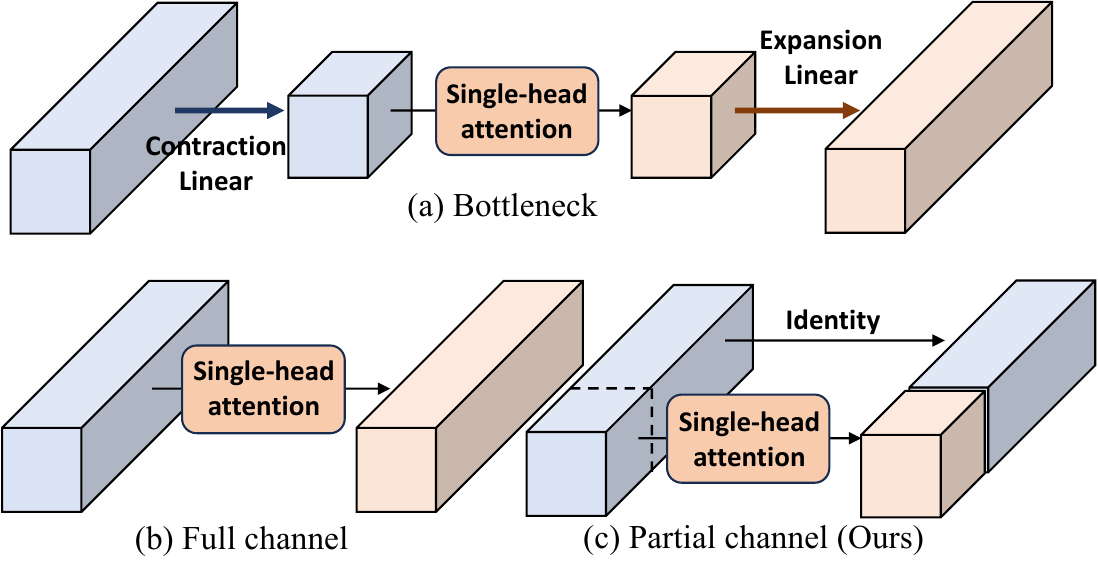}
    %\vspace{-20pt}

    % 테이블 부분
    \resizebox{1.0\linewidth}{!}{%
    \begin{tabular}{lC{2cm}C{2cm}C{2cm}C{3cm}C{2cm}}
    \hline
    Single-Head Attention design & \; Params \newline (M) & CPU$_{\text{ONNX}}$  \newline \text{(images/s)} & \; GPU \newline (images/s) & \; \; Top-1 \newline (\%) \\
    \hline
    (a) : Bottleneck & 10.6 & 802 & 27195 & 73.6 \\
    (b) : Full channel& 10.4 & 724 & 26432 & 75.1  \\
    \rowcolor{m}
    (c) : Partial channel (Ours) & 11.4 & \textbf{951}  & \textbf{26878}  & \textbf{75.2} \\
    \hline
    \end{tabular} }
    \caption{\textbf{Comparison of Single-head attention designs.} \textbf{(a)} replaces convolution with single-head attention in ResNet's bottleneck block~\cite{he2016resnet}. The contraction ratio is equal to the partial ratio in (c). \textbf{(b)} uses full channels for single-head attention modules. All models are configured to have similar speeds. Our partial channel approach has the best speed-accuracy tradeoff.}
    \label{fig: single_head_choices}
\end{figure}

In Fig.~\ref{fig: single_head_choices}, we also explore various single-head designs.
Recent studies~\cite{alternet, EMO, liu2023efficientvit, xiao2021early, wu2021cvt, guo2022cmt, vasu2023fastvit} sequentially combines convolution and attention layers to incorporate local details into a global contexts.
Unfortunately, this approach can only extract either local detail or global context in a given token mixer.
Also, it is noted in~\cite{raghu2021vitdocnn} that some channels process local details while others handle global modeling.
These observations imply that the current serial approaches have redundancy when processing all channels in each layer (Fig.~\ref{fig: single_head_choices} (a), (b)).
In contrast, our partial channel approach with preceding convolution memory-efficiently addresses the aforementioned issue by leveraging two complementary features in parallel within a single token mixer~\cite{alternet, inceptionformer}.

\begin{figure}[t]
    \centering
    
    % 이미지 부분
    %\includegraphics[width=1.0\linewidth]{Fig/runtime_profiling_5.pdf}
    \includegraphics[height=4cm,width=1.0\linewidth]{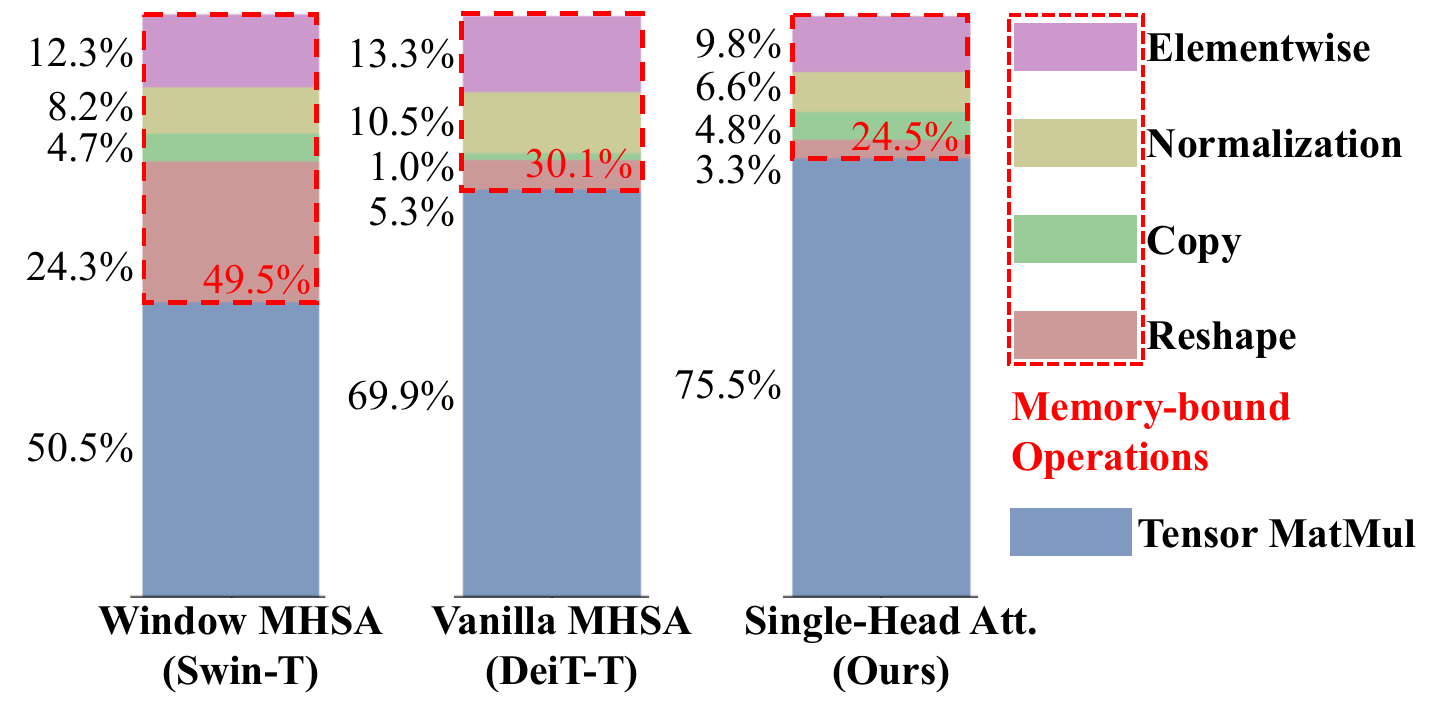}
    %\vspace{-20pt}
    
    \caption{\textbf{Runtime breakdown.} Operations highlighted in the red box represent memory-bound operations, where the majority of the duration is consumed by memory accesses, and the computational time is relatively brief.}
    \label{fig: runtime_profiling}
\end{figure}

For effective utilization of the attention layer, Layer Normalization~\cite{ba2016layernormalization} is essential; meanwhile, to implement a multi-head approach, data movements like reshape operation are required.
Consequently, as shown in Fig.~\ref{fig: runtime_profiling}, a large portion of MHSA's runtime is taken up by memory-bound operations like reshaping and normalization~\cite{liu2023efficientvit, ivanov2021data, tabani2021improving, dao2022flashattention}.
By minimizing the use of memory-bound operations or applying them to fewer input channels, the SHSA module can fully leverage the computing power of GPUs/CPUs.

\subsection{Single-Head Vision Transformer}\label{sec: shvit}
An overview of the Single-Head Vision Transformer (SHViT) architecture is illustrated in Fig.~\ref{fig:Model}.
Given an input image, we first apply four 3$\times$3 strided convolution layers to it.
Compared to the stride-16 16$\times$16 convolution tokenizing used by standard ViT models~\cite{vit, deit}, our overlapping patchify stem can extract better local representations~\cite{graham2021levit, xiao2021early, liu2023efficientvit}.
Then, the tokens pass through three stages of stacked SHViT blocks for hierarchical representation extraction.
A SHViT block consists of three main modules (see Fig.~\ref{fig:Model}): Depthwise Convolution (DWConv) layer for local feature aggregation or conditional position embedding~\cite{conditional, chu2021twins}, Single-Head Self-Attention (SHSA) layer for modeling global contexts, and Feed-Forward-Network (FFN) for channel interaction.
The expansion ratios in FFN are set to 2.
The combination of DWConv and SHSA captures both local and global dependencies in a computationally and memory-efficient manner.
Based on findings in sec.~\ref{sec: micro_design}, we do not use the SHSA layer in the first stage.
To reduce tokens without information loss, we utilize an efficient downsampling layer, which is composed of two stage 1 blocks, with an inverted residual block~\cite{mobilenetv2, mobilenetv3, liu2023efficientvit} (stride-2) placed between them.
Finally, the global average pooling and fully connected layer are used to output the predictions.

Besides the aforementioned operators, normalization and activation layers also play crucial roles in determining the model speed.
We employ Layer Normalization~\cite{ba2016layernormalization} only for the SHSA layer while integrating Batch Normalization (BN)~\cite{ioffe2015batch} into the remaining layers, as BN can be merged into its adjacent convolution or linear layers.
We also use ReLU~\cite{nair2010relu} activations instead of other complex alternatives~\cite{hendrycks2016gelu, chen2020dynamicrelu, mobilenetv3}, as they are much slower on various inference deployment platforms~\cite{vasu2023mobileone, vasu2023fastvit, liu2023efficientvit}.

We build four SHViT variants with different settings of depth and width.
Due to the large-sized patch embedding and single-head design, we can use a larger number of channels and blocks than previous efficient models.
Model specifications are provided in Tab.~\ref{tab:variants}.

\definecolor{m}{RGB}{230, 230, 230}
\begin{table}[t]
\resizebox{1.0\linewidth}{!}{%
\renewcommand{\arraystretch}{1.0}
\begin{tabular}{l|ccc|c|c}
\hline
\multicolumn{1}{c|}{Model variants} & Depth         & Emb. dim.           & Reso. & Partial ratio             & Exp. ratio         \\ \hline
SHViT-S1                            & {[}2, 4, 5{]} & {[}128, 224, 320{]} & 224   & \multirow{4}{*}{1 / 4.67} & \multirow{4}{*}{2} \\
SHViT-S2                            & {[}2, 4, 5{]} & {[}128, 308, 448{]} & 224   &                           &                    \\
SHViT-S3                            & {[}3, 5, 5{]} & {[}192, 352, 448{]} & 224   &                           &                    \\
SHViT-S4                            & {[}4, 7, 6{]} & {[}224, 336, 448{]} & 256   &                           &                    \\ \hline
\end{tabular}
}
\caption{Architecture details of SHViT variants.}
\label{tab:variants}
\end{table}

\section{Experiments}\label{experiments}

\subsection{Implementation Details}
We conduct image classification on ImageNet-1K~\cite{deng2009imagenet}, which includes 1.28M training and 50K validation images for 1000 categories.
All models are trained from scratch using AdamW~\cite{loshchilov2017adamw} optimizer for 300 epochs with a learning rate of 10$^{-3}$ and a total batch size of 2048.
We use cosine learning rate scheduler~\cite{loshchilov2016sgdr} with linear warmup for 5 epochs.
Weight decays are set to 0.025/0.032/0.035/0.03 for SHViT-S1 to S4.
For fair comparison, we follow the same data augmentation proposed in ~\cite{deit}, including Mixup~\cite{zhang2017mixup}, random erasing~\cite{zhong2020randomerase}, and auto-augmentation~\cite{cubuk2019autoaugment}.
For 384$^2$ and 512$^2$ resolution, we finetune the model for 30 epochs with weight decay of 10$^{-8}$ and learning rate of 0.004. 
Additionally, we assess throughput performance across various hardware platforms.
We measure GPU throughput on an Nvidia A100 with batch size of 256.
For CPU and $\text{CPU}_{\text{ONNX}}$, we evaluate the runtime on an Intel(R) Xeon(R) Gold 5218R CPU @ 2.10GHz processor , with batch size of 16 (using a single thread).
For $\text{CPU}_{\text{ONNX}}$, we convert the models to ONNX~\cite{onnx} runtime format.
Mobile latency is measured using iPhone 12 with iOS version 16.5.
We export the models (batch size is set to 1) using CoreML tools~\cite{coreml2017} and report the median latency over 1,000 runs.
We also validate our model as an efficient vision backbone for object detection and instance segmentation on COCO~\cite{lin2014microsoftcoco} with RetinaNet~\cite{lin2017focal} and Mask R-CNN~\cite{he2017mask}, respectively.
All models are trained under 1$\times$ schedule (12 epochs) following ~\cite{liu2021swin} on mmdetection library~\cite{mmdetection}.

\definecolor{m}{RGB}{230, 230, 230}
% \usepackage{multirow}
\begin{table*}[h!]
\centering
\resizebox{0.7\linewidth}{!}{%
\renewcommand{\arraystretch}{1.0}
\begin{tabular}{cccccccccc}
\hline
\multicolumn{1}{c}{\multirow{2}{*}{Model}} & \multirow{2}{*}{Reso.} & \multirow{2}{*}{Epochs} & FLOPs & \multicolumn{1}{c}{Params} & \multicolumn{3}{c}{Throughput (images/s)} & Top-1 & Top-5 \\ \cline{6-8}
\multicolumn{1}{c}{}                       &                        &                         & (M)   & \multicolumn{1}{c}{(M)}    & GPU    & CPU  & \multicolumn{1}{c}{$\text{CPU}_{\text{ONNX}}$} & (\%)  & (\%)  \\ \hline
MobileNetV3-Small~\cite{mobilenetv3}                           & 224                    & 600                     & 57    & 2.5                         & 31477  & 167  & 1172                      & 67.4  & -     \\
MobileViT-XXS~\cite{mehta2022mobilevit}                               & 256                    & 300                     & 410   & 1.3                         & 7594   & 21   & 170                       & 69.0  & -     \\
MobileViTV2 ×0.5~\cite{mehta2022mobilevitv2}                            & 256                    & 300                     & 466   & 1.4                         & 8616   & 17   & 157                       & 70.2  & -     \\
EfficientViT-M2~\cite{liu2023efficientvit}                             & 224                    & 300                     & 201   & 4.2                         & 30377  & 147  & 781                       & 70.8  & 90.2  \\
MobileOne-S0~\cite{vasu2023mobileone}                                & 224                    & 300                     & 275   & 2.1                         & 19689  & 86   & \textbf{1648}                      & 71.4  & -     \\
EMO-1M~\cite{EMO}                                      & 224                    & 300                     & 261   & 1.3                         & 10032  & 34   & 119                       & 71.5  & -     \\
FasterNet-T0~\cite{chen2023fasternet}                                & 224                    & 300                     & 340   & 3.9                         & 23518  & 92   & 844                       & 71.9  & -     \\
ShuffleNetV2 ×1.5~\cite{shufflenetv2}                           & 224                    & 300                     & 299   & 3.5                         & 16495  & 62   & 799                       & 72.6  & -     \\
MobileFormer-96M~\cite{mobileformer}                            & 224                    & 450                     & 96    & 3.6                         & 13106  & 91   & 235                       & \textbf{72.8}  & -     \\ \rowcolor{m}
\textbf{SHViT-S1}                                    & 224                    & 300                     & 241   & 6.3                         & \textbf{33489}  & 143  & 1111                      & \textbf{72.8}  & 91.0  \\ \hline
EfficientFormerV2-S0~\cite{li2022efficientformerv2}                        & 224                    & 300                     & 400   & 3.5                         & 2374   & 54   & 372                       & 73.7  & -     \\
EfficientViT-M4~\cite{liu2023efficientvit}                             & 224                    & 300                     & 299   & 8.8                         & 26201  & 113  & 616                       & 74.3  & 91.8  \\
EdgeViT-XXS~\cite{pan2022edgevits}                                 & 224                    & 300                     & 600   & 4.1                         & 6763   & 33   & 168                       & 74.4  & -     \\
MobileViT-XS~\cite{mehta2022mobilevit}                                & 256                    & 300                     & 986   & 2.3                         & 4408   & 8    & 96                        & 74.8  & -     \\
ShuffleNetV2 ×2.0~\cite{shufflenetv2}                           & 224                    & 300                     & 591   & 7.4                         & 12276  & 40   & 250                       & 74.9  & 92.4  \\
EMO-2M~\cite{EMO}                                      & 224                    & 300                     & 439   & 2.3                         & 7333   & 25   & 78                        & 75.1  & -     \\
MobileNetV3-Large~\cite{mobilenetv3}                           & 224                    & 600                     & 217   & 5.4                         & 13994  & 43   & 613                       & \textbf{75.2}  & -     \\ \rowcolor{m}
\textbf{SHViT-S2}                                    & 224                    & 300                     & 366   & 11.4                        & \textbf{26878}  & 99   & \textbf{951}                       & \textbf{75.2}  & 92.4  \\ \hline
FastViT-T8~\cite{vasu2023fastvit}                                  & 256                    & 300                     & 700   & 2.1                         & 5978   & 23   & 140                       & 75.6  & -     \\
GhostNet ×1.3~\cite{ghostnet}                               & 224                    & 300                     & 226   & 7.3                         & 9433   & 39   & 109                       & 75.7  & 92.7  \\
FasterNet-T1~\cite{chen2023fasternet}                                & 224                    & 300                     & 850   & 7.6                         & 17827  & 41   & 552                       & 76.2  & -     \\
EfficientNet-B0~\cite{efficientnet}                             & 224                    & 350                     & 390   & 5.3                         & 8433   & 26   & 267                       & 77.1  & 93.3  \\
EfficientViT-M5~\cite{liu2023efficientvit}                             & 224                    & 300                     & 522   & 12.4                        & 18722  & 64   & 456                       & 77.1  & 93.4  \\
PoolFormer-S12~\cite{yu2022metaformer}                              & 224                    & 300                     & 1823  & 11.9                        & 5432   & 13   & 120                       & 77.2  & -     \\
MobileOne-S2~\cite{vasu2023mobileone}                                & 224                    & 300                     & 1299  & 7.8                         & 9355   & 22   & 581                       & \textbf{77.4}  & -     \\ \rowcolor{m}
\textbf{SHViT-S3}                                    & 224                    & 300                     & 601   & 14.2                        & \textbf{20522}  & 62   & \textbf{731}                       & \textbf{77.4}  & 93.4  \\ \hline
EdgeViT-XS~\cite{pan2022edgevits}                                  & 224                    & 300                     & 1100  & 6.7                         & 5520   & 21   & 120                       & 77.5  & -     \\
EfficientFormerV2-S1~\cite{li2022efficientformerv2}                        & 224                    & 300                     & 650   & 6.1                         & 2112   & 37   & 325                       & 77.9  & -     \\
MobileViTV2 ×1.0~\cite{mehta2022mobilevitv2}                            & 256                    & 300                     & 1800  & 4.9                         & 4345   & 7    & 63                        & 78.1  & -     \\
ResNet50~\cite{liu2022convnext, he2016resnet}                            & 224                    & 300                     & 4110  & 25.6                         & 5281   & 8    & 271                        & 78.8  & -     \\
FasterNet-T2~\cite{chen2023fasternet}                                & 224                    & 300                     & 1910  & 15.0                        & 11181  & 21   & 417                       & 78.9  & -     \\
EMO-6M~\cite{EMO}                                      & 224                    & 300                     & 961   & 6.1                         & 5105   & 15   & 50                        & 79.0  & -     \\
EfficientNet-B1~\cite{efficientnet}                             & 240                    & 350                     & 700   & 7.8                         & 4982   & 11   & 156                       & 79.1  & 94.4  \\
FastViT-T12~\cite{vasu2023fastvit}                                 & 256                    & 300                     & 1400  & 6.8                         & 4197   & 14   & 92                        & 79.1  & -     \\
MobileFormer-508M~\cite{mobileformer}                           & 224                    & 450                     & 508   & 14.0                        & 5390   & 23   & 91                        & 79.3  & -     \\
MobileOne-S4~\cite{vasu2023mobileone}                                & 224                    & 300                     & 2978  & 14.8                        & 5281   & 11   & 281                       & \textbf{79.4}  & -     \\ \rowcolor{m}
\textbf{SHViT-S4}                                    & 256                    & 300                     & 986   & 16.5                        & \textbf{14283}  & 36   & \textbf{509}                       & \textbf{79.4}  & 94.5  \\ \hline
\multicolumn{10}{c}{Finetuning with higher resolution}                                                                                                                                             \\ \hline
$\text{EfficientViT-M5}_{r384}$~\cite{liu2023efficientvit}                             & \textbf{384}                    & 330                     & 1486  & 12.4                        & 7041   & 17   & 176                       & 79.8  & 95.0  \\
$\text{EfficientViT-M5}_{r512}$~\cite{liu2023efficientvit}                             & \textbf{512}                    & 360                     & 2670  & 12.4                        & 3777   & 9    & 88                        & 80.8  & 95.5  \\ \rowcolor{m}
\textbf{$\text{SHViT-S4}_{r384}$}                                    & \textbf{384}                    & 330                     & 2225  & 16.5                        & \textbf{6702}   & 14   & \textbf{315}                       & \textbf{81.0}  & 95.4  \\ \rowcolor{m}
\textbf{$\text{SHViT-S4}_{r512}$}                                    & \textbf{512}                    & 360                     & 3973  & 16.5                        & \textbf{3957}   & 8    & \textbf{198}                       & \textbf{82.0}  & 95.9  \\ \hline
\end{tabular}}
\vspace{-2 mm}
\caption{SHViT classification performance on ImageNet-1K~\cite{deng2009imagenet} with comparisons to SOTA efficient models. Throughput is measured on an Nvidia A100 GPU with batch size of 256 for GPU and Intel(R) Xeon(R) Gold 5218R CPU @ 2.10GHz processor with batch size of 16 for CPU and CPU$_{\text{ONNX}}$.
Larger throughput means faster inference speed. FLOP count is computed by fvcore~\cite{fvcore} library.}
\label{tab: imagenet_results}
\end{table*}
\subsection{SHViT on ImageNet-1K Classification}
As shown in Fig.~\ref{fig: acc_fps_plot}, Tab.~\ref{tab: imagenet_results}, and ~\ref{table:mobile_latency}, we compare Single-Head Vision transformer (SHViT) with the state-of-the-art models.
The comparison results clearly show that our SHViT achieves a better trade-off between accuracy and throughput/latency across various devices.

\noindent\textbf{Comparison with efficient CNNs.} Our SHViT-S1 achieves 5.4\% higher accuracy than MobileNetV3-Small~\cite{mobilenetv3}, while maintaining similar speeds on both the A100 GPU and Intel CPU.
Compared to ShuffleNetV2 $\times$2.0~\cite{shufflenetv2}, SHViT-S2 obtains slightly better performance with 2.2$\times$ and 2.5$\times$ speed improvements on the A100 GPU and Intel CPU, respectively.
Furthermore, our model is 3.8$\times$ faster when converted to the ONNX runtime format.
Compared to the recent FasterNet-T1~\cite{chen2023fasternet}, SHViT-S3 not only achieves 1.2\% higher accuracy but also runs faster: 15.1\% on the A100 GPU and 32.4\% on the Intel CPU.
%SHViT-S4 also achieves slightly better performance  with the searched model EfficientNet-B1~\cite{efficientnet}, while runs 2.9$\times$/3.3$\times$ faster on the A100 GPU/Intel CPU with ONNX implementation.
Notably, at Top-1 accuracy of 79.1$-$79.4\%, our model is 2.9$\times$/ 3.3$\times$ faster than EfficientNet-B1~\cite{efficientnet}, 3.4$\times$/ 5.5$\times$ faster than FastViT-T12~\cite{vasu2023fastvit}, and 2.7$\times$/ 1.8$\times$ faster than MobileOne-S4~\cite{vasu2023mobileone} on the A100 GPU/Intel CPU with ONNX format.
%These results show that 
\textit{When leveraging minimal attention module in a memory-efficient way, ViTs can still show fast inference speeds like efficient CNNs.}

\noindent\textbf{Comparison with efficient ViTs and hybrid models.} Our SHViT-S1 has 10\% and 42\% higher throughput than EfficientViT-M2~\cite{liu2023efficientvit} on the A100 GPU and Intel CPU with ONNX runtime, respectively, while showing better performance with a large margin (70.8\% $\rightarrow$ 72.8\%).
%Also, SHViT-S3 runs more 3.8$\times$ and 6.1$\times$ faster than PoolFormer-S12~\cite{yu2022metaformer} with slightly improved score.
SHViT-S3 obtains similar accuracy to PoolFormer-S12~\cite{yu2022metaformer}, but it uses 3$\times$ fewer FLOPs, is 3.8$\times$ faster on the A100 GPU, and 6.1$\times$ faster as ONNX model.
Remarkably, our SHViT-S4 surpasses recent EdgeViT-XS~\cite{pan2022edgevits} with a 1.9\% higher accuracy, while being 2.6$\times$ faster on A100 GPU, 1.7$\times$ faster on Intel CPU, and 4.2$\times$ faster on ONNX implementation. 
As shown in the results above, when converted to ONNX format, our models demonstrate a notable performance boost compared to the recent SOTA models.
This enhancement is largely because our single-head design uses fewer reshape operations, which often cause overhead in ONNX runtime.
\textit{To summarize, the above results demonstrate that our proposed memory-efficient macro design has a more significant impact on the speed-accuracy tradeoff than efficient attention variants or highly simple operations like pooling.}

\definecolor{m}{RGB}{230, 230, 230}
\begin{table}[t]
\resizebox{1.0\linewidth}{!}{%
\renewcommand{\arraystretch}{0.9}
\begin{tabular}{lcccccc}
\hline
\multicolumn{1}{c}{\multirow{2}{*}{Model}} & \multirow{2}{*}{Reso.} & \multicolumn{1}{c}{Flops} & \multicolumn{3}{c}{Throughput (images/s)} & Top-1 \\ \cline{4-6}
\multicolumn{1}{c}{}                       &                        & \multicolumn{1}{c}{(M)}   & GPU    & CPU  & \multicolumn{1}{c}{$\text{CPU}_{\text{ONNX}}$} & (\%)  \\ \hline \rowcolor{m}
\textbf{SHViT-S1}                                    & 224                    & 241                        & \textbf{33489}  & 143  & \textbf{1111}                      & \textbf{74.0}  \\ \hline
SwiftFormer-XS~\cite{shaker2023swiftformer}                              & 224                    & 600                        & 7922   & 26   & 175                       & 75.7  \\
EfficientFormerV2-S0~\cite{li2022efficientformerv2}                        & 224                    & 400                        & 2374   & 54   & 372                       & 75.7  \\ \rowcolor{m}
\textbf{SHViT-S2}                                    & 224                    & 366                        & \textbf{26878}  & 99   & \textbf{951}                       & \textbf{76.2}  \\ \hline
FastViT-T8~\cite{vasu2023fastvit}                                  & 256                    & 1400                       & 5978   & 23   & 140                       & 76.7  \\ \rowcolor{m}
\textbf{SHViT-S3}                                    & 224                    & 601                        & \textbf{20522}  & 62   & \textbf{731}                       & \textbf{78.3}  \\ \hline
SwiftFormer-S~\cite{shaker2023swiftformer}                               & 224                    & 1000                       & 6415   & 21   & 147                       & 78.5  \\
EfficientFormerV2-S1~\cite{li2022efficientformerv2}                        & 224                    & 650                        & 2112   & 37   & 325                       & 79.0  \\
EfficientFormer-L1~\cite{li2022efficientformer}                          & 224                    & 1300                       & 6840   & 21   & 274                       & 79.2  \\ \rowcolor{m}
\textbf{SHViT-S4}                                    & 256                    & 986                        & \textbf{14283}  & 36   & \textbf{509}                       & \textbf{80.2}  \\ \hline
\end{tabular}}
\vspace{-2mm}
\caption{Comparison of SOTA efficient models on ImageNet-1K classification, using DeiT~\cite{deit} distillation recipe.}
\label{tab: distillation}
\end{table} 

\noindent\textbf{Finetuning with higher resolution.} Following~\cite{liu2023efficientvit}, we also finetune our SHViT-S4 to higher resolutions.
Compared to the state-of-the-art $\text{EfficientViT-M5}_{r512}$~\cite{liu2023efficientvit}, our $\text{SHViT-S4}_{r384}$ attains competitive performance, even when trained at a lower resolution. Additionally, $\text{SHViT-S4}_{r384}$ is 77.4\% faster on the A100 GPU, 55.6\% on the Intel CPU, and an impressive 3.6$\times$ faster on ONNX runtime format.
Moreover, $\text{SHViT-S4}_{r512}$ achieves 82.0\% top-1 accuracy with throughput of 3957 images/s on the A100 GPU, demonstrating effectiveness across various input sizes.

\noindent\textbf{Distillation results.} We report the performance of our models using DeiT~\cite{deit} distillation recipe in Tab.~\ref{tab: distillation}.
Notably, our models outperform competing models in both speed and accuracy.
Specifically, SHViT-S3 even surpasses FastViT-T8 which is 5.2$\times$ slower as ONNX models.
SHViT-S4 attains superior performance than EfficientFormer-L1~\cite{li2022efficientformer} while being 2.1$\times$ / 1.9$\times$ faster on the GPU / ONNX runtime.

\begin{figure}[t]
    \centering
    
    %\vspace{-5mm}
    \includegraphics[width=1.0\linewidth]{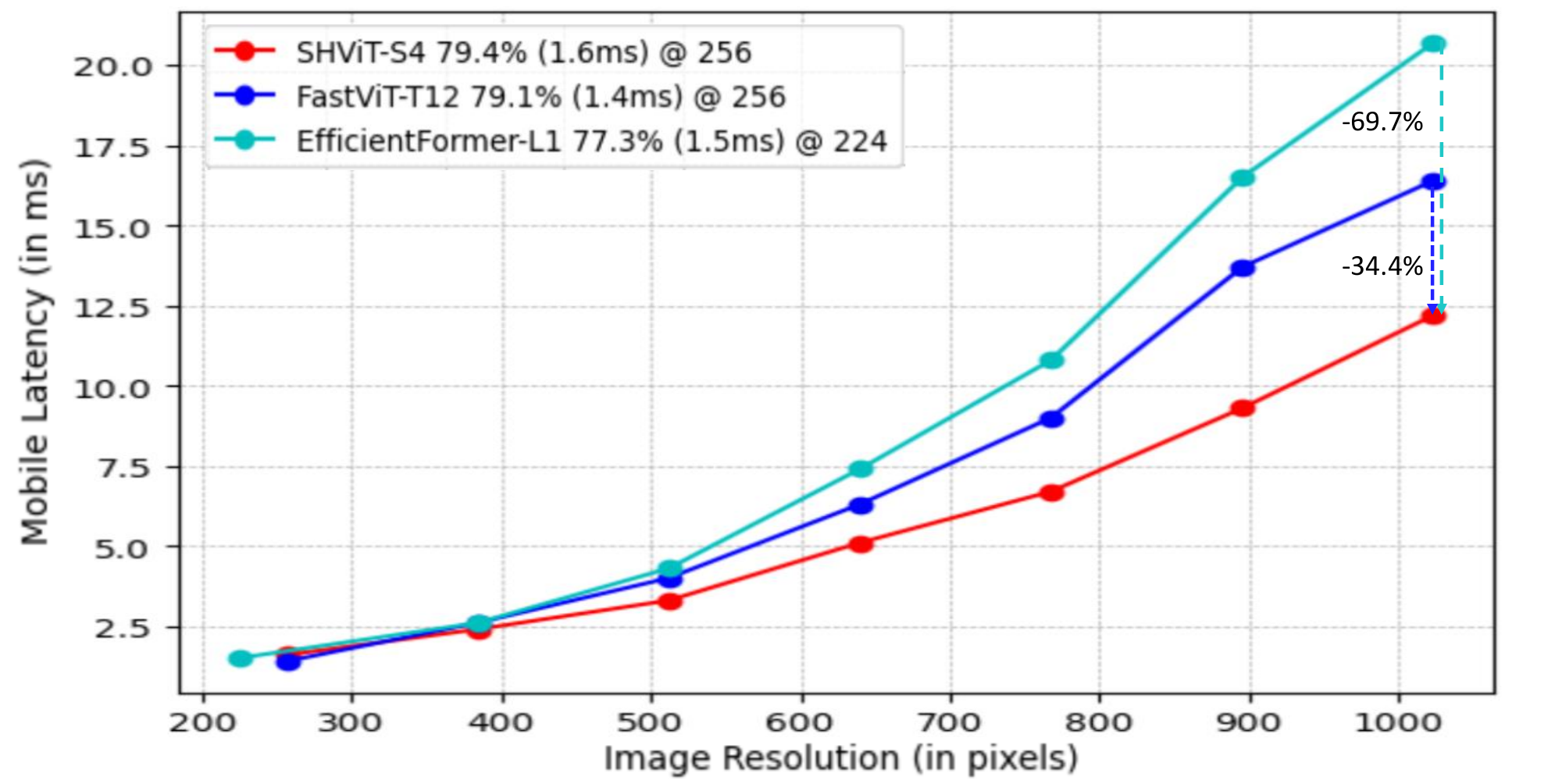}
    %\vspace{-4mm}
    \caption{Mobile latency comparison of a SHViT-S4 with recent state-of-the-art FastViT~\cite{vasu2023fastvit} and EfficientFormer~\cite{li2022efficientformer}; measured on iPhone 12 for various input sizes.}
    \label{fig: mobile_reso}
\end{figure}

\begin{wraptable}{l}{0.23\textwidth}
%\centering
\vspace{-0.3cm}
\tiny
\renewcommand{\arraystretch}{1.15}
\begin{tabular}{lcl}
\hline
\multicolumn{1}{l}{\multirow{2}{*}{Model}} & Latency & \multicolumn{1}{c}{Top-1} \\
\multicolumn{1}{c}{}                       & (ms)           & \multicolumn{1}{c}{(\%)}  \\ \hline
EfficientFormer-L1~\cite{li2022efficientformer}                               & 1.5            & 77.3 (79.2)               \\
EfficientFormerV2-S1~\cite{li2022efficientformerv2}                              & 1.3            & 77.9 (79.0)               \\
MobileViTv2 $\times$1.0~\cite{mehta2022mobilevitv2}                               & 3.8            & 78.1                      \\
EfficientNet-B1~\cite{efficientnet}                                   & 1.8            & 79.1                      \\
FastViT-T12~\cite{vasu2023fastvit}                                 & 1.4            & 79.1 (80.3)               \\
MobileOne-S4 ~\cite{vasu2023mobileone}                                  & 1.7            & 79.4                      \\ \rowcolor{m}
SHViT-S4                                   & 1.6            & 79.4 (80.2)               \\ \hline
\end{tabular}
\vspace{-0.1cm}
\caption{Mobile latency comparison. The results in brackets are trained with distillation~\cite{deit}.}
\label{table:mobile_latency}
\end{wraptable}
%\vspace{0.5cm}
\noindent\textbf{Mobile Latency Evaluation.} We also verify the effectiveness of our model on the mobile device in Tab.~\ref{table:mobile_latency}.
Compared to the efficient models EfficientNet-B1~\cite{efficientnet} / MobileOne-S4~\cite{vasu2023mobileone}, our SHViT-S4 achieves similar accuracy while running 0.2 ms / 0.1 ms faster on iPhone 12 device. 
SHViT-S4 also obtains competitive performance against highly-optimized models for mobile latency, indicating its consistent performance across diverse inference platforms.
Further results in Fig.~\ref{fig: mobile_reso} show that our model significantly outperforms over the recent models FastViT~\cite{vasu2023fastvit} and EfficientFormer~\cite{li2022efficientformer}, especially at higher resolutions.
At low resolutions, SHViT-S4 is slightly slower, but at 1024 $\times$ 1024, our model achieves 34.4\% and 69.7\% lower latency than FastViT and EfficientFormer, respectively.
These results stem from the increased memory efficiency in the macro and micro design.

\definecolor{m}{RGB}{230, 230, 230}
% \usepackage{multirow}
\begin{table}[t]
\resizebox{1.0\linewidth}{!}{%
\renewcommand{\arraystretch}{1.15}
\begin{tabular}{cccccccccc}
\hline
\multicolumn{10}{c}{RetinaNet Object Detection on COCO}                                                                                                                                                                               \\ \hline
\multicolumn{1}{c|}{\multirow{2}{*}{Backbone}} & \multicolumn{3}{c|}{Latency (ms)}         & \multirow{2}{*}{AP} & \multirow{2}{*}{$\text{AP}_{50}$} & \multirow{2}{*}{AP$_{75}$} & \multirow{2}{*}{AP$_s$} & \multirow{2}{*}{AP$_m$} & \multirow{2}{*}{AP$_l$} \\ \cline{2-4}
\multicolumn{1}{c|}{}                          & GPU  & CPU  & \multicolumn{1}{c|}{Mobile} &                     &                       &                       &                      &                      &                      \\ \hline
\multicolumn{1}{l|}{MobileNetV3~\cite{mobilenetv3}}               & 0.34 & 7.5  & \multicolumn{1}{c|}{7.5}   & 29.9                & 49.3                  & 30.8                  & 14.9                 & 33.3                 & 41.1                 \\
\multicolumn{1}{l|}{EfficientViT-M4~\cite{liu2023efficientvit}}           & 0.33 & 7.3  & \multicolumn{1}{c|}{7.8}    & 32.7                & 52.2                  & 34.1                  & 17.6                 & 35.3                 & 46.0                 \\
\multicolumn{1}{l|}{PVTv2-B0~\cite{wang2022pvtv2}}         & 0.73 & 115.4 & \multicolumn{1}{c|}{27.5}   & 37.2                & 57.2                  & 39.5                  & \textbf{23.1}                 & 40.4                 & 49.7                 \\
\multicolumn{1}{l|}{MobileFormer-508M~\cite{mobileformer}}         & 0.89 & 35.7 & \multicolumn{1}{c|}{26.9}   & 38.0                & 58.3                  & 40.3                  & 22.9                 & 41.2                 & 49.7                 \\ \multicolumn{1}{l|}{EdgeViT-XXS~\cite{pan2022edgevits}}         & 0.88 & 38.4 & \multicolumn{1}{c|}{12.9}   & 38.7                & 59.0                  & 41.0                  & 22.4                 & 42.0                 & 51.6                 \\\rowcolor{m}
\multicolumn{1}{l|}{\textbf{SHViT-S4}}                  & \textbf{0.28} & \textbf{5.0}  & \multicolumn{1}{c|}{\textbf{3.3}}    & \textbf{38.8}                & \textbf{59.8}                  & \textbf{41.1}                  & 22.0                 & \textbf{42.4}                 & \textbf{52.7}                 \\ \hline
\multicolumn{10}{c}{Mask R-CNN Object Detection \& Instance Segmentation on COCO}                                                                                                                                                     \\ \hline
\multicolumn{1}{c|}{\multirow{2}{*}{Backbone}} & \multicolumn{3}{c|}{Latency (ms)}         & \multirow{2}{*}{AP$^b$}  & \multirow{2}{*}{AP$^b_{50}$}    & \multirow{2}{*}{AP$^b_{75}$}    & \multirow{2}{*}{AP$^m$}   & \multirow{2}{*}{AP$^m_{50}$}   & \multirow{2}{*}{AP$^m_{75}$}   \\ \cline{2-4}
\multicolumn{1}{c|}{}                          & GPU  & CPU  & \multicolumn{1}{c|}{Mobile} &                     &                       &                       &                      &                      &                      \\ \hline
\multicolumn{1}{l|}{EfficientNet-B0~\cite{efficientnet}}           & 0.54 & 16.7 & \multicolumn{1}{c|}{3.8}    & 31.9                & 51.0                  & 34.5                  & 29.4                 & 47.9                 & 31.2                 \\
\multicolumn{1}{l|}{EfficientViT-M4~\cite{liu2023efficientvit}}           & 0.33 & 7.3  & \multicolumn{1}{c|}{7.8}    & 32.8                & 54.4                  & 34.5                  & 31.0                 & 51.2                 & 32.2                 \\
\multicolumn{1}{l|}{PoolFormer-S12~\cite{yu2022metaformer}}            & 1.20 & 40.4 & \multicolumn{1}{c|}{6.8}    & 37.3                & 59.0                  & 40.1                  & 34.6                 & 55.8                 & 36.9                 \\
\multicolumn{1}{l|}{EfficientFormer-L1~\cite{li2022efficientformer}}        & 0.84 & 21.0 & \multicolumn{1}{c|}{4.3}    & 37.9                & 60.3                  & 41.0                  & 35.4                 & 57.3                 & 37.3                 \\
\multicolumn{1}{l|}{ResNet-50~\cite{he2016resnet}}                 & 0.94 & 19.0 & \multicolumn{1}{c|}{8.8}    & 38.0                & 58.6                  & 41.4                  & 34.4                 & 55.1                 & 36.7                 \\
\multicolumn{1}{l|}{FastViT-SA12~\cite{vasu2023fastvit}}              & 1.06 & 39.4 & \multicolumn{1}{c|}{6.5}    & 38.9                & 60.5                  & \textbf{42.2}                  & \textbf{35.9}                 & 57.6                 & \textbf{38.1}                 \\ \rowcolor{m}
\multicolumn{1}{l|}{\textbf{SHViT-S4}}                  & \textbf{0.28} & \textbf{5.0}  & \multicolumn{1}{c|}{\textbf{3.3}}    & \textbf{39.0}                & \textbf{61.2}                  & 41.9                  & \textbf{35.9}                 & \textbf{57.9}                 & 37.9                 \\ \hline
\end{tabular}}
\vspace{-2mm}
\caption{Comparison results on object detection and instance segmentation on COCO 2017~\cite{lin2014microsoftcoco} using RetinaNet~\cite{lin2017focal} and Mask RCNN~\cite{he2017mask} head.
Backbone latencies are measured with image crops of 512 $\times$ 512. The batch sizes used for GPU, CPU, and Mobile latency are 32, 16, and 1 respectively.}
\label{tab: downstream_results}
\end{table}
\subsection{SHViT on downstream tasks}
In Tab.~\ref{tab: downstream_results}, We evaluate the transfer ability of our SHViT using two frameworks: 1) RetinaNet~\cite{lin2017focal} for object detection, 2) Mask R-CNN~\cite{he2017mask} for instance segmentation.

\noindent\textbf{Object detection.} SHViT-S4 is 2.3$\times$ faster on mobile device than MobileNetV3~\cite{mobilenetv3} and outperforms it by +8.9 AP.
Compared to MobileFormer~\cite{mobileformer}, our model achieves better performance while being 3.2$\times$ and 8.2$\times$ faster on the A100 GPU and mobile device, respectively.

\noindent\textbf{Instance Segmentation.} SHViT-S4 surpasses GPU or mobile-optimized models like EfficientViT~\cite{liu2023efficientvit} and EfficientNet~\cite{efficientnet} in speed, while delivering a substantial performance boost.
Remarkably, our model gains 1.7 $\text{AP}^{b}$ and 1.3 $\text{AP}^{m}$ over PoolFormer~\cite{yu2022metaformer} but runs 4.3$\times$, 8.1$\times$, and 2.1$\times$ faster on the GPU, CPU, and mobile device respectively.

As shown in the above results, the large-stride patchify stem with 3-stage reduces not only computational costs but also generates meaningful token representations, especially at higher resolutions. 
Furthermore, the marked performance gap with EfficientViT~\cite{liu2023efficientvit}, using a similar macro design, proves the efficacy of our micro design choices.

% \usepackage{multirow}
\definecolor{m}{RGB}{230, 230, 230}
\begin{table}[]
\resizebox{1.0\linewidth}{!}{%
\renewcommand{\arraystretch}{1.1}
\begin{tabular}{cclcccc}
\hline
\multirow{2}{*}{\#Row} & \multirow{2}{*}{Ablation}      & \multicolumn{1}{c}{\multirow{2}{*}{Variant}} & \multicolumn{3}{c}{Throughput (images/s)} & Top-1 \\ \cline{4-6}
                       &                                & \multicolumn{1}{c}{}                         & GPU         & CPU       & $\text{CPU}_{\text{ONNX}}$       & (\%)  \\ \hline
(1)                    & Single Head                 & $\rightarrow$ MHSA~\cite{vit, attention}     & 18036       & 50        & 578             & 77.7 \\
(2)                    & Self-Attention                    & $\rightarrow$ None                           & 22075       & 61        & 792             & 76.3  \\
\hline
(3)                    &                                & $= 1 / 8$                                        & 20666       & 57        & 754             & 77.1  \\ \rowcolor{m}
(4)                    &     Partial Ratio                           & $= 1 / 4.67$ (SHViT-S3)                               & 20522       & 62        & 731             & 77.4  \\
(5)                    &  & $= 1 / 2$                                        & 19976       & 56        & 673             & 77.5  \\ \hline
\end{tabular}}
\vspace{-2mm}
\caption{Ablation on our proposed Single-Head Attention and design choice for SHViT-S3 variant.}
\vspace{-5mm}
\label{tab: ablation}
\end{table}

\subsection{Ablation Study}\label{ablations}
In this section, we first verify the effectiveness of our proposed Single-Head Self-Attention (SHSA) layer and then conduct a concise ablation study on the value of the partial ratio for SHSA layer.
Results are provided in Tab.~\ref{tab: ablation}.

\noindent\textbf{Effectiveness of SHSA.} To assess whether the SHSA layer can effectively capture the global contexts like the Multi-Head Self-Attention (MHSA)~\cite{vit} layer, we conduct an ablation study by either replacing the SHSA layer with the MHSA layer or removing it.
As shown in Tab.~\ref{tab: ablation} (1, 2 \textit{vs.} 4), SHSA layer exhibits a better speed-accuracy tradeoff compared to MHSA layer. 
While removing SHSA layer results in a faster model speed, it leads to a significant drop in accuracy.
Meanwhile, model (2) can also achieve highly competitive performance compared with the SOTA models in Tab.~\ref{tab: imagenet_results}, which shows that our proposed macro design offers a solid architectural baseline under tight latency constraints.

\noindent\textbf{Searching for the appropriate partial ratio of SHSA.} By default, we set the partial ratio to 1 / 4.67 for all SHViT models, which obtains the optimal speed-accuracy tradeoff (3, 5 \textit{vs.} 4). 
Compared to a very small value, increasing the channels moderately for token interaction achieves effective performance enhancement at low costs.
Also, a too large value does not provide a performance boost that compensates for the accompanying costs.

\section{Related Work}
Leveraging Convolutional Neural Networks (CNN) in resource-constrained devices has gained significant attention from many researchers.
Within this trend, several strategies have emerged, including decomposition of convolution in MobileNets~\cite{mobilenet1, mobilenetv2, mobilenetv3}, channel shuffling in ShuffleNets~\cite{shufflenet, shufflenetv2}, cheap linear transformation in GhostNet~\cite{ghostnet}, compound scaling law in EfficientNet~\cite{efficientnet}, and structural re-parameterization in many works~\cite{ding2021repvgg, vasu2023mobileone, vasu2023fastvit}.

Even within the Vision Transformer (ViT)~\cite{vit} realm, there are ongoing numerous efforts for efficient designs to accelerate inference speed on various devices.
One promising approach is designing a new ViT architecture that integrates the local priors of CNN.
%And this also can be categorized into two streams: serial and parallel.
This method mostly incorporates attention only in the latter stages, allowing for the efficient extraction of global information without considerable computational overhead~\cite{li2022efficientformer, li2022efficientformerv2, alternet, graham2021levit, vasu2023fastvit}.
In contrast, other methods employ attention and convolution in parallel, either within a single token mixer~\cite{inceptionformer, pan2022hilo, xu2021vitae} or on a block-by-block basis~\cite{mobileformer}, to combine a rich set of features.
Another line of approach focuses on reducing the computational complexity of MHSA~\cite{pan2022edgevits, mehta2022mobilevit, mehta2022mobilevitv2, shaker2023swiftformer}.
For example, MobileViTv2~\cite{mehta2022mobilevitv2} introduces a separable self-attention with linear complexity with respect to the number of tokens (resolution).
EdgeViT~\cite{pan2022edgevits} applies MHSA to sub-sampled features to perform approximately full spatial interaction in a cost-effective manner.
%Our proposed solution addresses the two aspects overlooked by the previous works and simultaneously boosts memory efficiency.
Unlike the above approaches, we prioritize organizing tokens with minimal spatial redundancy over efficiently mixing tokens.

Also, recent works~\cite{michel2019sixteen, voita2019analyzing, chen2021svite, song2022cpvit, hou2022multicompression, yang2023vitprune, liu2021singlehead, liu2023efficientvit, chen2022principle, xia2023budgeted} have demonstrated that numerous heads function in similar ways and can be pruned without notably affecting performance.
EfficientViT~\cite{liu2023efficientvit} proposes feeding attention heads with
different splits of the full channel to  improve attention diversity.
In addition, ~\cite{chen2022principle} presents a regularization loss for multi-head similarity, while \cite{zhou2021deepvit} explores head similarity across different layers.
As opposed to reducing multi-head redundancy, we design module with single-head configuration, which not only inherently prevents multi-head redundancy but also saves computation costs.

\setlength{\parskip}{0pt}
\section{Conclusion} \label{conclusion}
In this work, we have investigated redundancies at both the spatial and channel dimensions of the architectural design commonly used by many established models.
We then proposed 16$\times$16 patch embeddings with 3-scale hierarchical representations and Single-Head Self-Attention to address the computational redundancies.
We further present our versatile SHViT, built upon our proposed macro/micro designs, that achieves ultra-fast inference speed and high performance on diverse devices and vision tasks. 

\noindent\textbf{Discussion.} While our macro design is effective, there is a need for fine-grained (high-resolution) features to enhance performance further or to recognize small objects. 
Therefore, our future work focuses on devising cost-effective ways to utilize them. Integrating the single-head design into existing sophisticated attention methods will be an intriguing way to explore.

\section{Acknowledgement}
This work was supported by the National Research Foundation of Korea (NRF) grant funded by the Korea government (MSIT) (NO.RS-2022-00166109) and (2022M3J6A1084845). 

% for Seokju Yun.
% Also, this work was supported by the 2022 Research Fund of the University of Seoul for Youngmin Ro.
\small
\bibliographystyle{unsrt}
\bibliography{refs}
\twocolumn[
\centering
\Large
\textbf{SHViT: Single-Head Vision Transformer with Memory Efficient Macro Design}

\textbf{------ Supplementary Material ------} \\
\vspace{1.0em}

\appendix

]

This supplementary material presents additional comparison results, memory analysis, detection results, and experimental settings.

\section{Comparison with Tiny Variants of Large-scale Models }~\label{appendix:tiny}
We compare our model against tiny variants of established models in Tab.~\ref{tab: tiny}.
Our model, when applied to higher resolutions, outperforms state-of-the-art models in terms of parameter and throughput.
Compared to Swin-T~\cite{liu2021swin}, our $\text{SHViT-S4}_{r384}$ is 0.3\% inferior in accuracy but is 2.3$\times$ / 9.5$\times$ faster on the A100 GPU / Intel CPU.

\begin{figure}[t]
    \centering
    
    % 이미지 부분
    \includegraphics[height=3.7cm,width=1.0\linewidth]{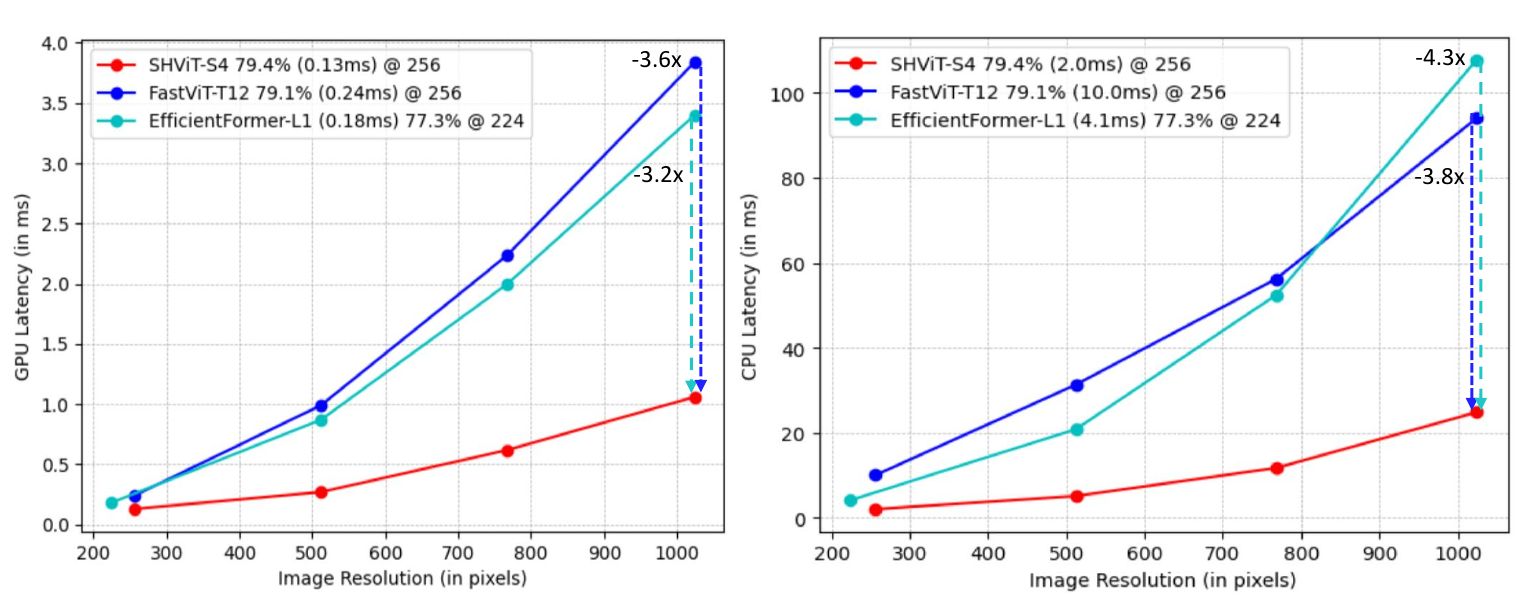}
    \vspace{-20pt}
    \caption{GPU, CPU latency comparison of a SHViT-S4 with recent state-of-the-art FastViT~\cite{vasu2023fastvit} and EfficientFormer~\cite{li2022efficientformer}; measured on A100 GPU, Intel CPU for various image resolutions.}
    \label{fig:image_res}
\end{figure}

In Fig.~\ref{fig:image_res}, we also provide further results of Section 3.2.
It demonstrates improved speed performance when increasing the resolution not only on mobile devices but also on other inference platforms compared to the recent models~\cite{li2022efficientformer, vasu2023fastvit}. This result showcases that our model can be a competitive alternative in real-world applications.
Further analysis of these performance enhancements will be detailed in the following section.

\section{Memory Efficiency Analysis}~\label{sec:appendix_memory}
Our model has a larger number of parameters compared to lightweight models.
For instance, SHViT-S3 has 2.7$\times$ more parameters compared to EfficientNet-B0~\cite{efficientnet}.
However, an important consideration for deploying the model on resource-constrained devices is the memory access cost of the feature maps.
On an I/O bound devices, the number of memory access for a given layer is as follows
\begin{gather}
    2 \times b \times h \times w \times c + k^2 \times c^2
\end{gather}
Particularly, when increasing batch size to enhance throughput, or for applications that require high-resolution input, the impact of the first term in the above equation becomes significantly more critical.
Our proposed macro and micro designs considerably reduce memory usage by eliminating redundancies in the first term's $h \times w$ and $c$ components, respectively. 
In Tab.~\ref{tab: memory}, our model, despite having more parameters than EfficientNet-B0, consumes less test memory. Notably, the disparity in memory usage grows with increasing batch sizes.

\section{Further Results on COCO Detection}~\label{sec:detr}
We also present results on COCO object detection benchmark~\cite{lin2014coco} with DEtection TRansformer (DETR)~\cite{carion2020detr, zhu2020deformable} framework in Tab.~\ref{tab: detr}.
\definecolor{m}{RGB}{230, 230, 230}

\begin{table}[t]
\resizebox{1.0\linewidth}{!}{%
\renewcommand{\arraystretch}{1.15}
\begin{tabular}{lccccc}
\hline
\multirow{2}{*}{Model} & Params & FLOPs & \multicolumn{2}{c}{Throughput (image/s)} & Top-1 \\ \cline{4-5}
                       & (M)    & (G)   & GPU                 & $\text{CPU}_{\text{ONNX}}$               & (\%)  \\ \hline
CaiT-XXS36~\cite{cait}             & 17     & 3.8   & 1394                & 24                 & 79.1  \\
Twins-PCPVT-S~\cite{chu2021twins}          & 24     & 3.8   & 3800                & 53                 & 81.2  \\
Swin-T~\cite{liu2021swin}                 & 28     & 4.5   & 2868                & 33                 & 81.3  \\ 
TNT-S~\cite{tnt}                  & 24     & 5.2   & 1554                & 37                 & 81.5  \\
CoAtNet-0~\cite{dai2021coatnet}              & 25     & 4.2   & 2448                & 53                 & 81.6  \\
DeiT-B~\cite{deit}                 & 87     & 17.6  & 3227                & 21                 & 81.8  \\
XCiT-S12~\cite{ali2021xcit}               & 26     & 4.8   & 3110                & -                  & 82.0  \\
PVTv2-B2~\cite{wang2022pvtv2}               & 25     & 4.0   & 2924                & 14                 & 82.0  \\
FocalNet-T~\cite{yang2022focal}             & 28     & 4.4   & 2808                & 68                 & 82.1  \\
ConvNeXt-T~\cite{liu2022convnet}             & 29     & 4.5   & 3325                & 49                 & 82.1  \\ \rowcolor{m}
\textbf{SHViT-S4}               & 17     & 1.0   & 14283               & 509                & 79.4  \\ \rowcolor{m}
\textbf{$\text{SHViT-S4}_{r384}$}              & 17     & 2.2   & 6702                & 315                & 81.0  \\ \rowcolor{m}
\textbf{$\text{SHViT-S4}_{r512}$}               & 17     & 4.0   & 3957                & 198                & 82.0  \\ \hline
\end{tabular}}
%\vspace{-2mm}
\caption{Comparison with the tiny variants of state-of-the-art large-scale models on ImageNet-1K classification. `r384' means finetuned at 384$\times$384 resolution. Models which could not be reliably converted to ONNX format are annotated by `$-$'.}
\label{tab: tiny}
\end{table}
\begin{table}[t]
\large
\resizebox{1.0\linewidth}{!}{%
\renewcommand{\arraystretch}{1.3}
\begin{tabular}{ccccccc}
\hline
\multirow{2}{*}{Model} & Top-1 & Params & \multicolumn{4}{c}{Inference Memory (MB) / Throughput (images/s)} \\ \cline{4-7} 
                       & (\%)  & (M)    & bs1          & bs32          & bs256          & bs1024         \\ \hline
SHViT-S3               & 77.4  & 14.2   & 1855 / 147   & 1963 / 4691   & 2613 / 20522   & 5525 / 22309   \\
EfficientNet-B0        & 77.1  & 5.3    & 1931 / 175   & 2015 / 5427   & 3861 / 8433    & 10493 / 8706   \\ \hline
\end{tabular}}
%\vspace{-2mm}
\caption{Memory Consumption Comparison with EfficientNet-B0~\cite{efficientnet}. `bs32' indicates that test time memory consumption and throughput are measured at  batch size of 32.}
\label{tab: memory}
\end{table}
The encoder of DETR consists of self-attention and FFN, and the decoder consists of self-attention, cross-attention, and FFN.
To demonstrate the efficacy of our single-head attention module not only as a feature extractor but also as a detection head, we apply single-head design to the encdoer's self-attention and decoder's cross-attention layers.
These two layers involve significant computational costs, thus employing a single-head design can greatly enhance the model speed.
However, in the detection head, each of the attention weights localizes different extremities~\cite{meng2021conditional}, making it challenging to simply combine them into a single-head design.
Furthermore, we find that the multi-head design in both the initial layer and latter layers of the encoder/decoder is vital.
Thus, we employ single-head attention modules in the 2nd, 3rd, and 4th layers of each encoder/decoder.
To minimize performance degradation, we also increase the head dimension in the single-head module from 32 to 64.
We train our model using the training recipe of Deformable DETR~\cite{zhu2020deformable, carion2020detr}.
As shown in Tab.~\ref{tab: detr}, single-head module demonstrates reasonable performance as a detector head and is a competitive alternative for applications where inference speed is crucial.

\begin{table}[t]
\large
\resizebox{1.0\linewidth}{!}{%
\renewcommand{\arraystretch}{1.4}
\begin{tabular}{ccccccccc}
\hline
Method                        & Params & \multicolumn{1}{c}{FPS} & AP   & AP$_{50}$ & AP$_{75}$ & AP$_S$  & AP$_M$  & AP$_L$  \\ \hline
Deformable DETR w/ single-head & 37.1M  & \multicolumn{1}{l}{31.4 \small \textcolor{red}{(24\% $\uparrow$)}}            & 43.1 & 62.7 & 46.6 & 26.3 & 46.6 & 57.2 \\
Deformable DETR               & 40.0M  & 25.4                    & 43.8 & 62.6 & 47.7 & 26.4 & 47.1 & 58.0 \\ \hline
\end{tabular}}
%\vspace{-2mm}
\caption{Effectiveness of our Single-Head Attention module with Deformable DETR~\cite{zhu2020deformable} framework. Our method improves test speed by 24\% without significant performance degradation.}
\label{tab: detr}
\end{table}

\section{More Details on Redundancy Experiments}
\noindent In this section, we provide implementation details of section 2.2.

\noindent\textbf{head similarity analysis}. For each layer $i$, the average cosine
similarity value is computed as:
\begin{gather}
    HeadSim_i = \frac{1}{N_h(N_h - 1)}\sum_{j \neq k}^{} cos(head_j, head_k)
\end{gather}
where $N_h$ is the number of heads. Then, the value is averaged for all batches.

\noindent\textbf{head ablation study}. In order to perform head ablation experiments, we modify the formula for Multi-Head Self-Attention (MHSA):
\begin{gather}
    \mathrm{MHSA} = \mathrm{Concat}(\textcolor{red}{\delta_1}\mathrm{head}_1,...,\textcolor{red}{\delta_N}\mathrm{head}_N)W^O, \\
    \mathrm{head}_i = \mathrm{Attention}(\textbf{X}_iW^Q_i, \textbf{X}_iW^K_i, \textbf{X}_iW^V_i), \\
    \mathrm{Attention}(\textbf{q}, \textbf{k}, \textbf{v}) = \mathrm{Softmax}(\textbf{q}\textbf{k}^\mathsf{T} / \sqrt{d_{head}})\textbf{v},
\end{gather}
where the $\delta$ are mask vaiables with values in \{0, 1\}.
When all $\delta$ are equal to 1, the above layer is equivalent to the MHSA layer.
In order to ablate head $i$, we simply set $\delta_i = 0$.
We conduct experiments by selectively removing one or more attention heads from a given architecture during test time and assessing the resulting impact on accuracy.
\textit{And we report the best accuracy for each layer in the model, i.e. the accuracy achieved by reducing the entire layer to the single most important head.}

We further investigate head redundancy in DeiT-S-Distill~\cite{deit}, a vision transformer distilled with knowledge from ConvNets.
In the distilled model, we can also observe a significant computational redundancy among many heads in the latter stages. Additionally, in the early stages, where many heads operate similarly to convolution, there is a relatively substantial decline in performance.

\section{Further Discussions on Related Works}

\noindent \textbf{About Macro Design.} Our patch embedding scheme is similar to that of \cite{graham2021levit,  liu2023efficientvit}, but the derivation process takes place from a completely different perspective.
While~\cite{graham2021levit} indirectly determines the patch embedding size through experiment grafting ResNet and DeiT, our work, on the other hand, investigate redundancy from the beginning, analyzing it separately in terms of spatial and channel. 
This allows us to address not only the spatial redundancy in traditional patch embedding but also propose a SHSA module, in contrast to \cite{graham2021levit} which employs MHSA (at mobile, SHViT-S4 80.2\%/\textbf{1.6}ms \textit{vs.} LeViT-192 80.0\%/\textbf{28.0}ms).
\textit{To the best of our knowledge, none of existing works have analyzed the effects (speed, memory efficiency) of resolving spatial redundancy in diverse environments (devices, tasks).}

\begin{figure}[t]
    \centering
    
    % 이미지 부분
    \includegraphics[width=1.0\linewidth]{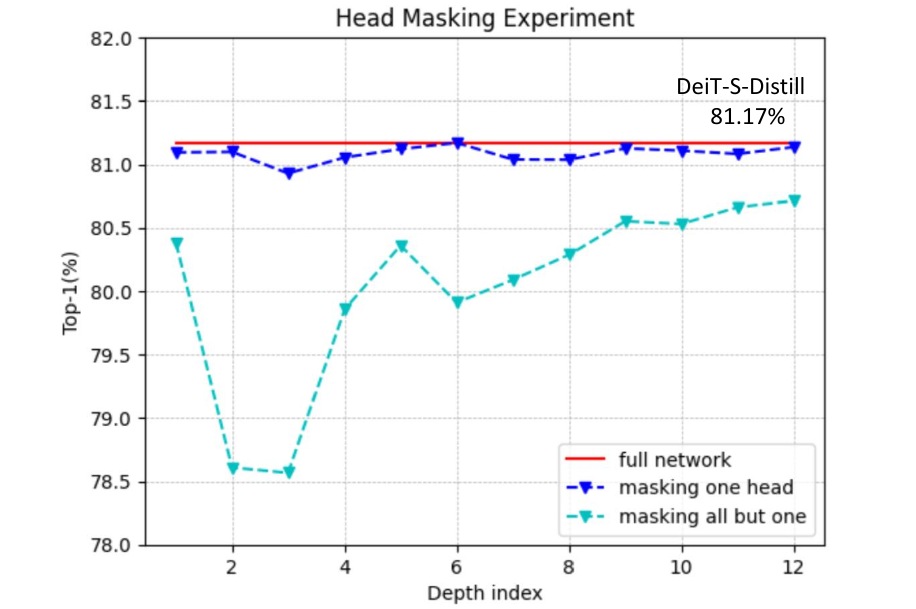}
    \vspace{-20pt}
    \caption{Head ablation study on DeiT-Small-Distill~\cite{deit}.}
    \label{fig:deit_redundancy}
\end{figure}

\noindent \textbf{About Partial Design in SHSA.} 
Partial channel design has also been employed in previous research~\cite{shufflenetv2, chen2023fasternet}. However, our work is distinct in both motivation and effectiveness. While prior work primarily focused on FLOPs (or throughput) and so employs convolutions (either depthwise or vanilla) on partial channels, this paper addresses multi-head redundancy by employing attention with single-head on partial channels. Furthermore, our SHSA, with preceding convolution, memory-efficiently leverages two complementary features in parallel within a single token mixer~\cite{alternet, inceptionformer}.

\end{document}